\documentclass{article} 
\usepackage{iclr2021_conference,times}

\usepackage{amsmath,amsfonts,bm}

\def\eqref#1{equation~\ref{#1}}

\def\1{\bm{1}}

\DeclareMathAlphabet{\mathsfit}{\encodingdefault}{\sfdefault}{m}{sl}
\SetMathAlphabet{\mathsfit}{bold}{\encodingdefault}{\sfdefault}{bx}{n}

\usepackage{multirow}
\usepackage{hyperref}
\usepackage{url}
\usepackage{makecell}
\usepackage{booktabs}
\usepackage{array}
\usepackage{comment}
\usepackage{soul}
\usepackage{graphicx}
\usepackage{caption}
\usepackage{subcaption}
\usepackage{microtype}
\usepackage{enumitem}
\usepackage{scrextend}
\usepackage{tabularx}
\newcommand{\norm}[1]{\left\lVert#1\right\rVert}
\newcolumntype{H}{>{\setbox0=\hbox\bgroup}c<{\egroup}@{}} 

\title{Network Pruning that Matters: \\ A Case Study on Retraining Variants}

\author{Duong H. Le\\
	VinAI Research, Vietnam\\
	\And
	Binh-Son Hua\\
	VinAI Research and VinUniversity, Vietnam\\
}

\newcommand{\duong}[1]{
	\textcolor{blue}{{#1}}
}

\newcommand{\best}[1]{
	\mathbf{\textcolor{blue}{{#1}}}
}

\newcommand{\secondbest}[1]{
	\textcolor{blue}{{#1}}
}

\iclrfinalcopy 
\begin{document}

	\maketitle
	
	\begin{abstract}
		Network pruning is an effective method to reduce the computational expense of over-parameterized neural networks for deployment on low-resource systems. 
		Recent state-of-the-art techniques for retraining pruned networks such as weight rewinding and learning rate rewinding have been shown to outperform the traditional fine-tuning technique in recovering the lost accuracy~\citep{renda2020comparing}, but so far it is unclear what accounts for such performance.  
		In this work, we conduct extensive experiments to verify and analyze the uncanny effectiveness of learning rate rewinding. We find that the reason behind the success of learning rate rewinding is the usage of a \emph{large} learning rate. Similar phenomenon can be observed in other learning rate schedules that involve large learning rates, e.g., the 1-cycle learning rate schedule \citep{smith2019super}. 
		By leveraging the \textit{right} learning rate schedule in retraining, we demonstrate a counter-intuitive phenomenon in that randomly pruned networks could even achieve better performance than methodically pruned networks (fine-tuned with the conventional approach). 
		Our results emphasize the cruciality of the learning rate schedule in pruned network retraining -- a detail often overlooked by practioners during the implementation of network pruning. 
	\end{abstract}

	\section{Introduction}
	Training neural networks is an everyday task in the era of deep learning and artificial intelligence.
	Generally speaking, given data availability, large and cumbersome networks are often preferred as they have more capacity to exhibit good data generalization. In the literature, large networks are considered easier to train than small ones \citep{neyshabur2018towards, arora2018optimization, novak2018sensitivity, brutzkus2019larger}. Thus, many breakthroughs in deep learning are strongly correlated to increasingly complex and over-parameterized networks. 
	
	However, the use of large networks exacerbate the gap between research and practice since real-world applications usually require running neural networks in low-resource environments for numerous purposes: reducing memory, latency, energy consumption, etc. To adopt those networks to resource-constrained devices, network pruning \citep{lecun1990optimal, han2015learning, li2016pruning} is often exploited to remove dispensable weights, filters and other structures from neural networks. The goal of pruning is to reduce overall computational cost and memory footprint without inducing significant drop in performance of the network.
	
	
	A common approach to mitigating performance drop after pruning is retraining: we continue to train the pruned models for some more epochs. In this paper, we are interested in approaches based on learning rate schedules to control the retraining. A well-known practice is \emph{fine-tuning}, which aims to train the pruned model with a small fixed learning rate. More advanced learning rate schedules exist, which we generally refer to as \emph{retraining}. The retraining step is a critical part in implementing network pruning, but it has been largely overlooked and tend to vary in each implementation including differences in learning rate schedules, retraining budget, hyperparameter choices, etc. 
	
	Recently, \cite{renda2020comparing} proposed a state-of-the-art technique for retraining pruned networks namely \emph{learning rate rewinding} (LRW). Specifically, instead of fine-tuning the pruned networks with a fixed learning rate, usually the last learning rate from the original training schedule \citep{han2015learning, liu2018rethinking}, the authors suggested using the learning rate schedule from the previous $t$ epochs (i.e. rewinding). This seemingly subtle change in learning rate schedule led to an important result: LRW was shown to achieve comparable performance to more complex and computationally expensive pruning algorithms while only utilizing simple norm-based pruning. Unfortunately, the authors did not provide the analysis to justify the improvement. In general, it is intriguing to understand the importance of a learning rate schedule and how it affects the final performance of a pruned model.
	
	In this work, we study the behavior of pruned networks under different retraining settings.  
	We found that the efficacy from retraining with learning rate rewinding is rooted in the use of a \emph{large} learning rate, which helps pruned networks to converge faster after pruning. We demonstrate that the success of learning rate rewinding over fine-tuning is \emph{not} exclusive to the learning rate schedule coupling with the original training process. Retraining with a large learning rate could outperform fine-tuning even with some modest retraining, e.g., for a few epochs, and regardless of network compression ratio.
	
	
	We argue that retraining is of paramount importance to regain the performance in network pruning and should not be overlooked when comparing two pruning algorithms. This is evidenced by our extensive experiments: (1) \emph{randomly} pruned network can outperform methodically pruned network  with only (hyper-parameters free) modifications of the learning rate schedule in retraining, and (2) a simple baseline such as norm-based pruning can perform as well as  as other complex pruning methods by using a large learning rate restarting retraining schedule.
	
	The contributions of our work are as follows.
	\begin{itemize}[leftmargin=*]
		
		\item We document a thorough experiment on learning rate schedule for the retraining step in network pruning with different pruning configurations;
		
		\item We show that learning rate matters: pruned models retrained with a \emph{large} learning rate consistently outperform those trained by conventional fine-tuning regardless of specific learning rate schedules;
		
		\item We present a novel and counter-intuitive result achieved by solely applying large learning rate retraining: a \textbf{randomly} pruned network and a simple norm-based pruned network can perform as well as networks obtained from more sophisticated pruning algorithms. 
		
	\end{itemize}
	
	Given the significant impact of learning rate schedule in network pruning, we advocate the following practices: learning rate schedule should be considered as a critical part of retraining when designing pruning algorithms. Rigorous ablation studies with different retraining settings should be made for a fair comparison of pruning algorithms. 
	To facilitate reproducibility, we would release our implementation upon publication.
	
	
	\section{Preliminary and Methodology}
	\label{sec:methodology}
	Pruning is a common method to produce compact and high performance neural networks from their original large and cumbersome counterparts.We can categorize pruning approaches into \textbf{three} classes: \textit{Pruning after training} - which consists of three steps: training the original network to convergence, prune redundant weights based on some criteria, and retrain the pruned model to regain the performance loss due to pruning \citep{li2016pruning, han2015learning, luo2017thinet, ye2018rethinking, wen2016learning, he2017channel}; \textit{Pruning during training} - we update the ``pruning mask" while training the network from scratch, thus, allowing pruned neurons to be recovered \citep{zhu2017prune, kusupati2020soft, wortsman2019discovering, Lin2020Dynamic, he2019filter, he2018soft}; \textit{Pruning before training} -  Inspired by the Lottery Ticket Hypothesis \citep{frankle2018lottery}, some recent works try to find the sparsity mask at initialization and train the pruned network from scratch without changing the mask \citep{lee2018snip, tanaka2020pruning, Wang2020Picking}.
	
	In this work, we are mainly concerned with the first category i.e. pruning after training which has the largest body of work to our knowledge. Traditionally, the last step is referred to as \emph{fine-tuning}, i.e., continue to train the pruned model with a small learning rate obtained from the last epoch of the original model. This seemly subtle step is often overlooked when designing pruning algorithms. 
	
	Particularly, we found that the implementation of previous pruning algorithms have many notable differences in their retraining step: some employed a small value of learning rate (e.g. $0.001$ on ImageNet) to fine-tune the network  \citep{molchanov2016pruning, li2016pruning, han2015learning} for a small number of epochs, e.g., $20$ epochs in the work by \cite{li2016pruning}; some used a larger value of learning rate ($0.01$) with much longer retraining budgets, e.g., $60$, $100$ and $120$ epochs respectively on ImageNet \citep{zhuang2018discrimination, gao2020discrete, li2020eagleeye}; \cite{you2019gate, li2020eagleeye} respectively utilized 1-cycle  \citep{smith2019super} and cosine annealing learning rate schedule instead of conventional step-wise schedule.
	Despite such difference, the success of each pruning algorithm is only attributed to the pruning algorithm itself. This motivates us to ask the question: do details like learning rate schedule used for retraining matter?
	
	In this section, we strive to understand the behavior of pruned models under different retraining configurations and how they impact the final performance. Specifically, we conduct experiments with different retraining schedules on simple baselines such as $\ell_1$-norm filters pruning \citep{li2016pruning} and magnitude-based weights pruning \citep{han2015learning}. We show that the efficacy of several pruning algorithms can be boosted by simply modifying the learning rate schedule. More importantly, the performance gain by retraining can be remarkable: the accuracy loss can drop to zero and in some cases better accuracy than baseline models can be achieved. 
	
	To analyze the effect of retraining a pruned network, we based on learning rate rewinding~\citep{renda2020comparing} and experiment with different retraining settings. Although in the previous work, \citet{renda2020comparing} demonstrated the efficacy of learning rate rewinding across datasets and pruning criteria, there is a lack of understanding of the actual reason behind the success of this technique. Here we hypothesize that the initial pruned network is a suboptimal solution, staying in a local minima. Learning rate rewinding succeeds because it uses a large learning rate to encourage the pruned networks to converge to another, supposingly better, local minima. Our experiment setups are as follows.

	
	\textbf{Retraining techniques.} To verify this conjecture empirically, we conduct experiments with different learning rate schedules including learning rate rewinding~\citep{renda2020comparing} while varying pruning algorithms, network architectures and datasets. In this work, we consider the following retraining techniques:
	\begin{enumerate}[leftmargin=*]
		\item \textsc{Fine-tuning} (FT) Fine-tuning is the most common retraining techniques \citep{han2015learning, li2016pruning, liu2018rethinking}. In this approach, we continue train the pruned networks for $t$ epochs with the last (smallest) learning rate of original training.
		\item \textsc{Learning Rate Rewinding} (LRW) \cite{renda2020comparing} propose to reuse the learning rate schedule of the original training when retraining pruned networks. Specifically, when retraining for $t$ epochs, we reuse the learning rate schedule from the previous $t$ epochs, i.e., rewinding.
		\item \textsc{Scaled Learning Rate Restarting} (SLR): In this approach, we employ the learning rate schedule that is proportionally identical to the standard training. For example, the learning rate is dropped by a factor of $10\times$ at $50\%$ and $75\%$ of retraining epochs on CIFAR, which is akin to original training learning rate adjustment. The original learning rate schedule can be found in Appendix A. 
		\item \textsc{Cyclic Learning Rate Restarting} (CLR): Instead of using stepwise learning rate schedule as \textit{scaled learning rate restarting}, we leverage the 1-cycle \citep{smith2019super}, which is shown to give faster convergence speed than conventional approaches.
	\end{enumerate}

	\begin{figure}
		\centering
		\def\sc{0.245}
		\begin{subfigure}[b]{\sc\linewidth}
			\includegraphics[width=\linewidth]{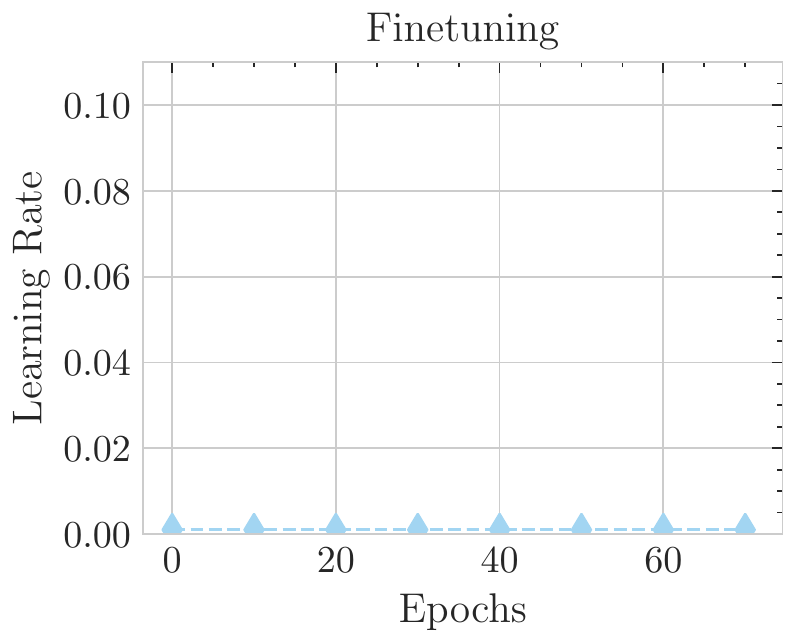}
			\caption{FT}
			\label{lrschedule:a}
		\end{subfigure}
		\begin{subfigure}[b]{\sc\linewidth}
			\includegraphics[width=\linewidth]{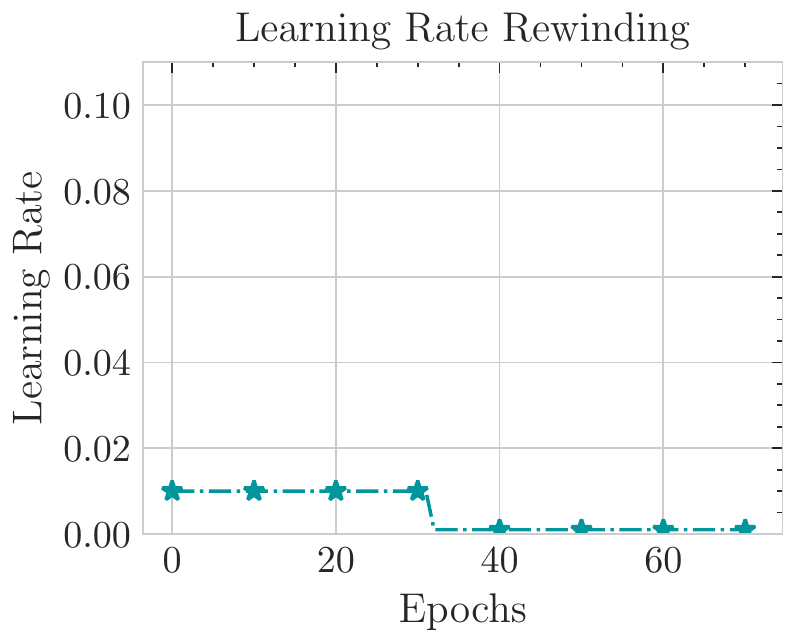}
			\caption{LRW}
			\label{lrschedule:b}
		\end{subfigure}
		\begin{subfigure}[b]{\sc\linewidth}
			\includegraphics[width=\linewidth]{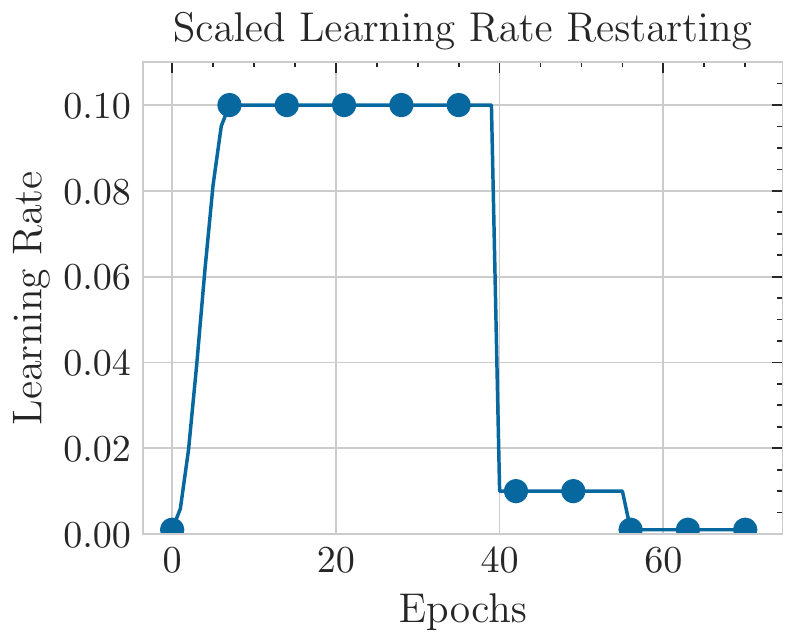}
			\caption{SLR}
			\label{lrschedule:c}
		\end{subfigure}
		\begin{subfigure}[b]{\sc\linewidth}
			\includegraphics[width=\linewidth]{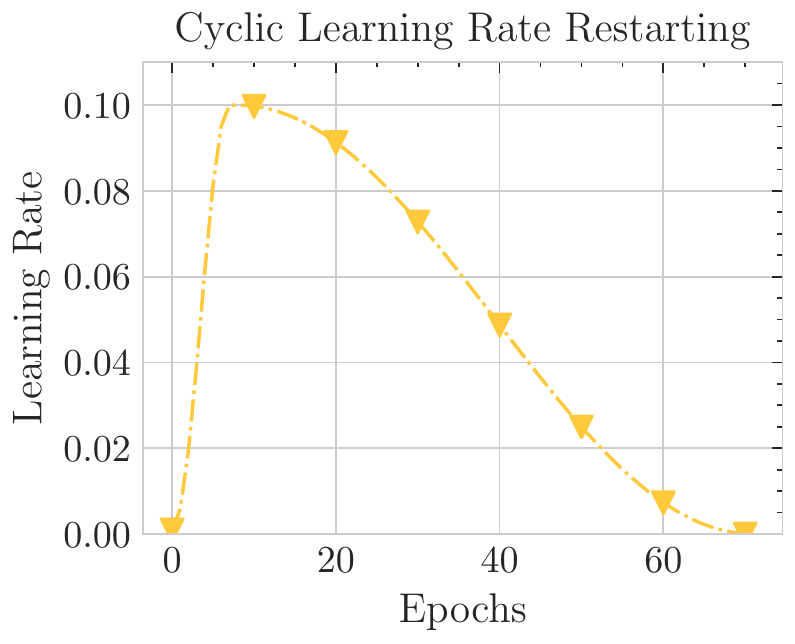}
			\caption{CLR}
			\label{lrschedule:d}
		\end{subfigure}
		\caption{
			Learning rate with different schedules on CIFAR when retraining for $72$ epochs. In (\subref{lrschedule:a}), the learning rate is fixed to the last learning rate of original training (i.e. $0.001$). In (\subref{lrschedule:b}), the learning rate is "rewound" to previous $72$ epochs (which is $0.01$), and is dropped to $0.001$ after $32$ epochs. In (\subref{lrschedule:c}), after warming up the learning rate, we drop its value by the factor of $10\times$ at $50\%$ and $75\%$ of remaining epochs. In (\subref{lrschedule:d}), we warm up the learning rate from the lowest to the highest value (of standard training) for the first few epochs, then decay the learning rate according to cosine function. 
		}
		\label{fig:lrschedule}
	\end{figure}

	Note that for the last two strategies (SLR and CLR), we \textit{warmup} the learning rate for $10\%$ of total retraining budget. For simplicity, we always use the \textit{largest} learning rate of the original training for \textit{learning rate restarting}. Specifically, the learning rate is increased from the smallest learning rate of original training to the largest one according to \textit{cosine} function. The detailed learning rate schedule of each technique is depicted in Figure \ref{fig:lrschedule}. See also Appendix~\ref{appendix:clr_warmup} for the choice of warmup epochs.
	
	\textbf{Pruning algorithms} We consider the following dimensions in our experiments. For pruning methods, we use {$\ell_1$-norm filters pruning} (PFEC) \citep{li2016pruning} and (global) magnitude-based weights pruning (\text{MWP}) \citep{han2015learning} and evaluate them on the CIFAR-10, CIFAR-100 and ImageNet dataset. We examine both variations of pruning namely \text{one-shot pruning} and \text{iterative pruning} when comparing the proposed retraining techniques. Furthermore, we also experiment the CLR schedule on HRank \citep{lin2020hrank}, Taylor Pruning (TP) \citep{molchanov2019importance} and Soft Filter Pruning (SFP) \citep{he2018soft}.
	
	Our implementation and hyperparameters of $\ell_1$-norm filters pruning and magnitude weight pruning are based on the public implementation of \cite{liu2018rethinking},
	which is shown to obtain comparable results with the original works. For remaining algorithms, we use official implementations with hyperparameters specified according to their papers. The detailed configurations of training and fine-tuning is provided in the Appendix \ref{sec:training_configuration} for interested readers.
	
	\textbf{Evaluation.} For  CIFAR-10 and CIFAR-100, we run each experiment three times and report ``$\mathrm{mean}\pm\mathrm{std}$". For ImageNet, we run each experiment once.  These settings are kept consistently across architectures, pruning algorithms, retraining techniques, and ablation studies unless otherwise stated.
	

	\section{A case study on Retraining in Network Pruning}
	\label{sec:case_study}
	
	\begin{figure}[t]
		\centering
		\def\sc{0.325}
		\begin{subfigure}[b]{\sc\textwidth}
			\centering
			\includegraphics[width=\textwidth]{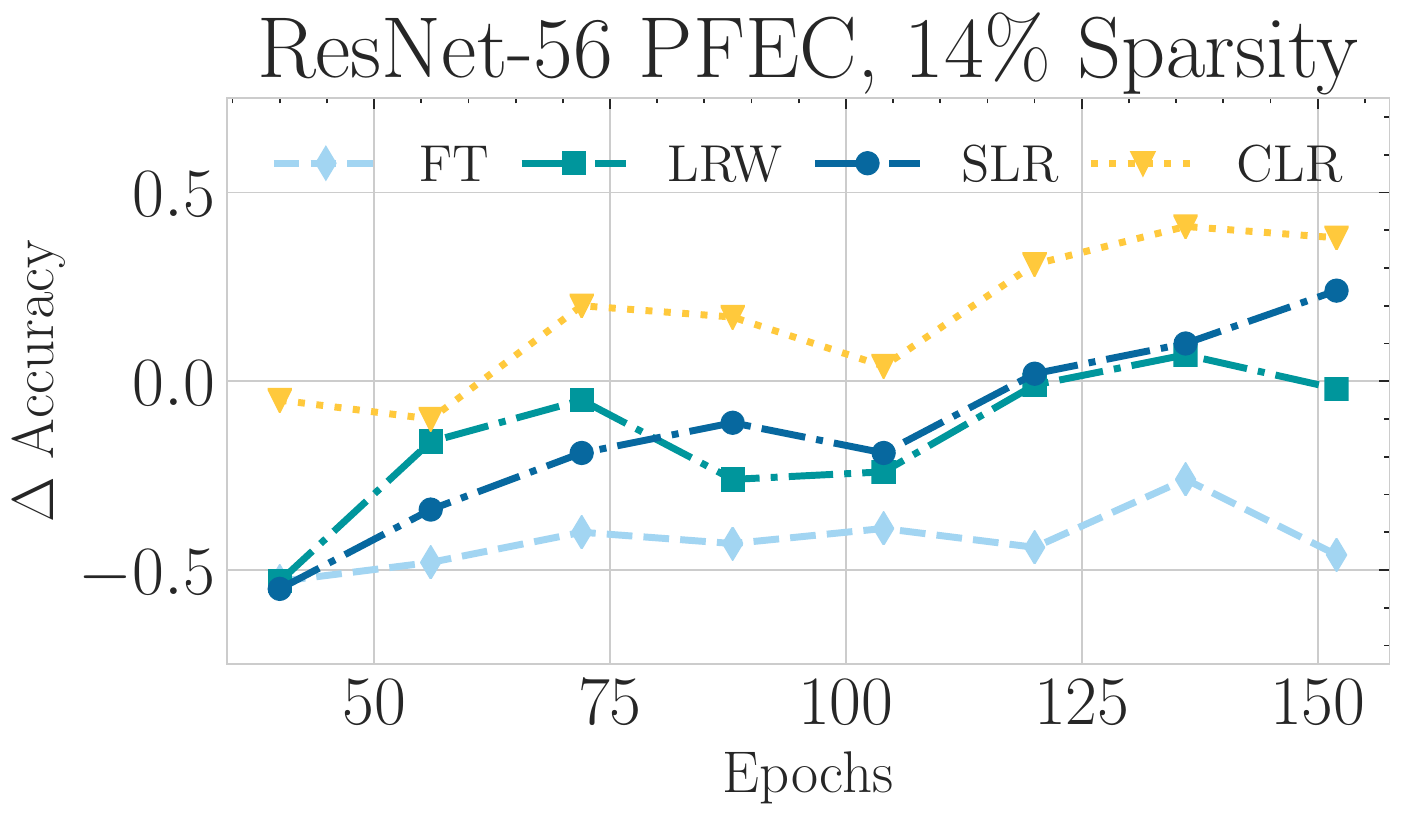}
			\label{fig:oneshot_pfec_cifar10_a}
		\end{subfigure}
		\hfill
		\begin{subfigure}[b]{\sc\textwidth}
			\centering
			\includegraphics[width=\textwidth]{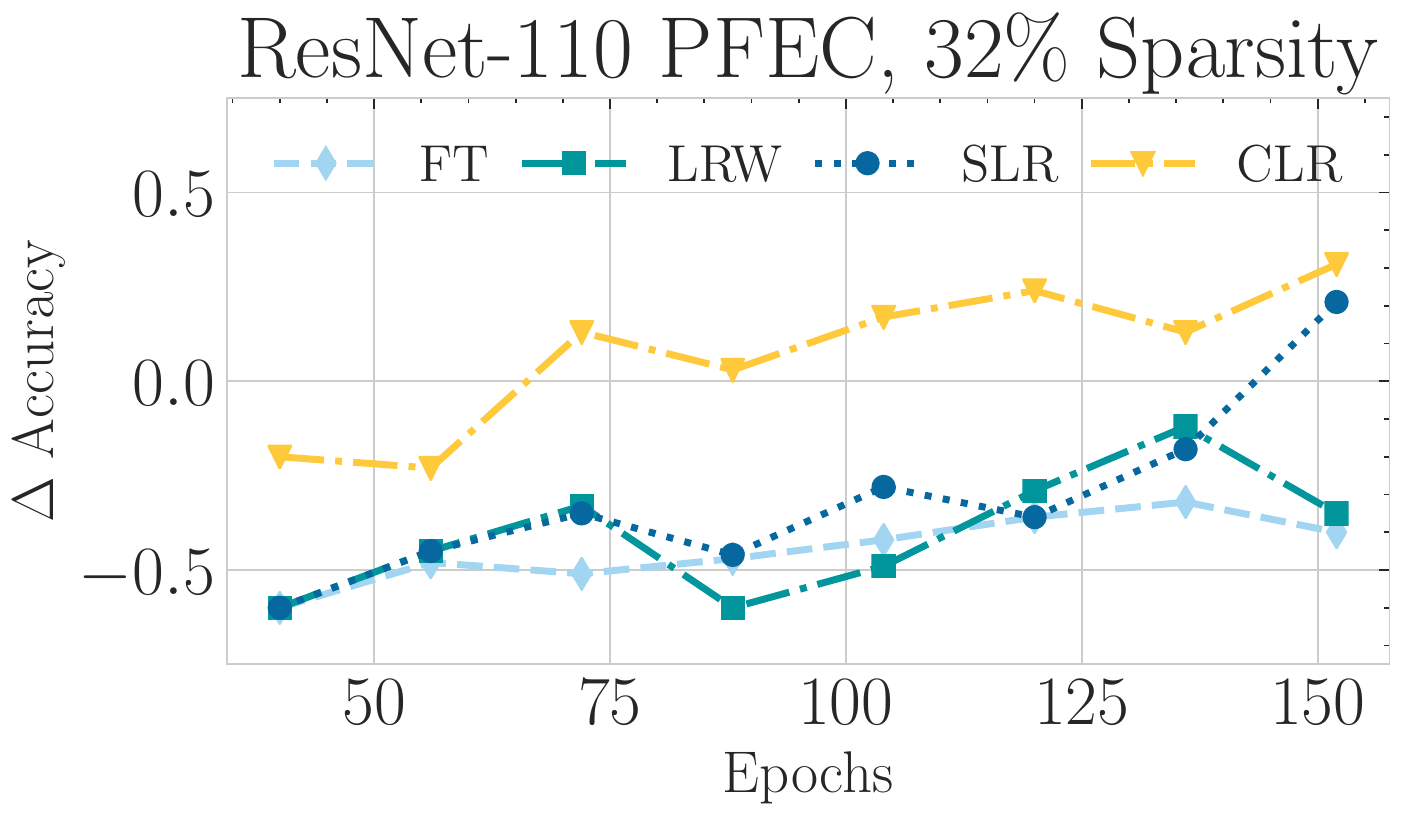}
			\label{fig:oneshot_pfec_cifar10_b}
		\end{subfigure}
		\hfill
		\begin{subfigure}[b]{\sc\textwidth}
			\centering
			\includegraphics[width=\textwidth]{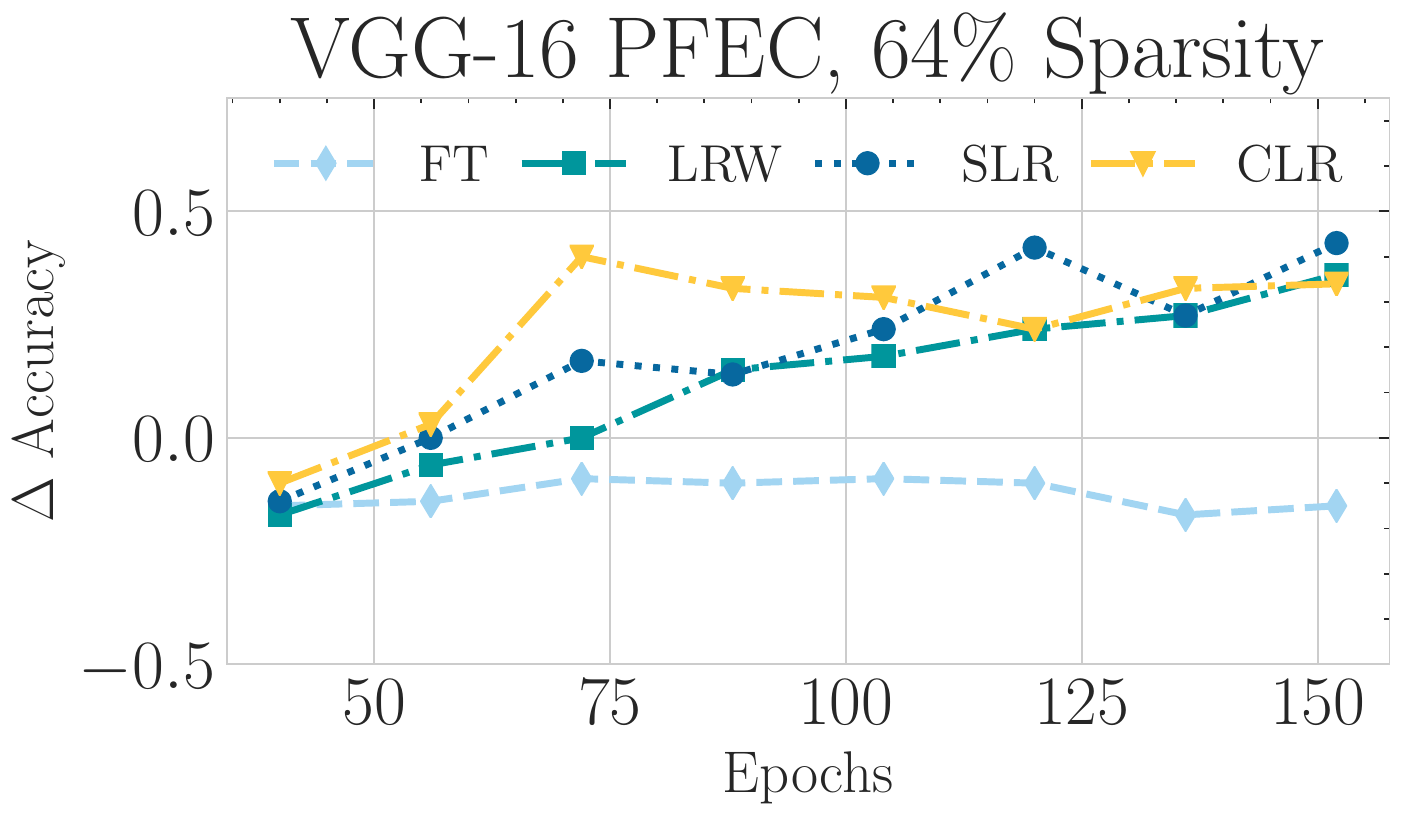}
			\label{fig:oneshot_pfec_cifar10_c}
		\end{subfigure}
		
		\caption{\textit{One-shot} \textit{structured} pruning on CIFAR-10 dataset using $\ell_1$-norm filters pruning \citep{li2016pruning} while varying retraining budgets. As can be seen, learning rate schedule matters. Schedules that employ large learning rates (LRW, SLR, CLR) are significantly better than fine-tuning. Among them, CLR performs best in most cases.}
		\label{fig:oneshot_pfec_cifar10_original}
	\end{figure}
	
	\begin{figure}[t]
		\centering
		\def\sc{0.325}
		\begin{subfigure}[b]{\sc\textwidth}
			\centering
			\includegraphics[width=\textwidth]{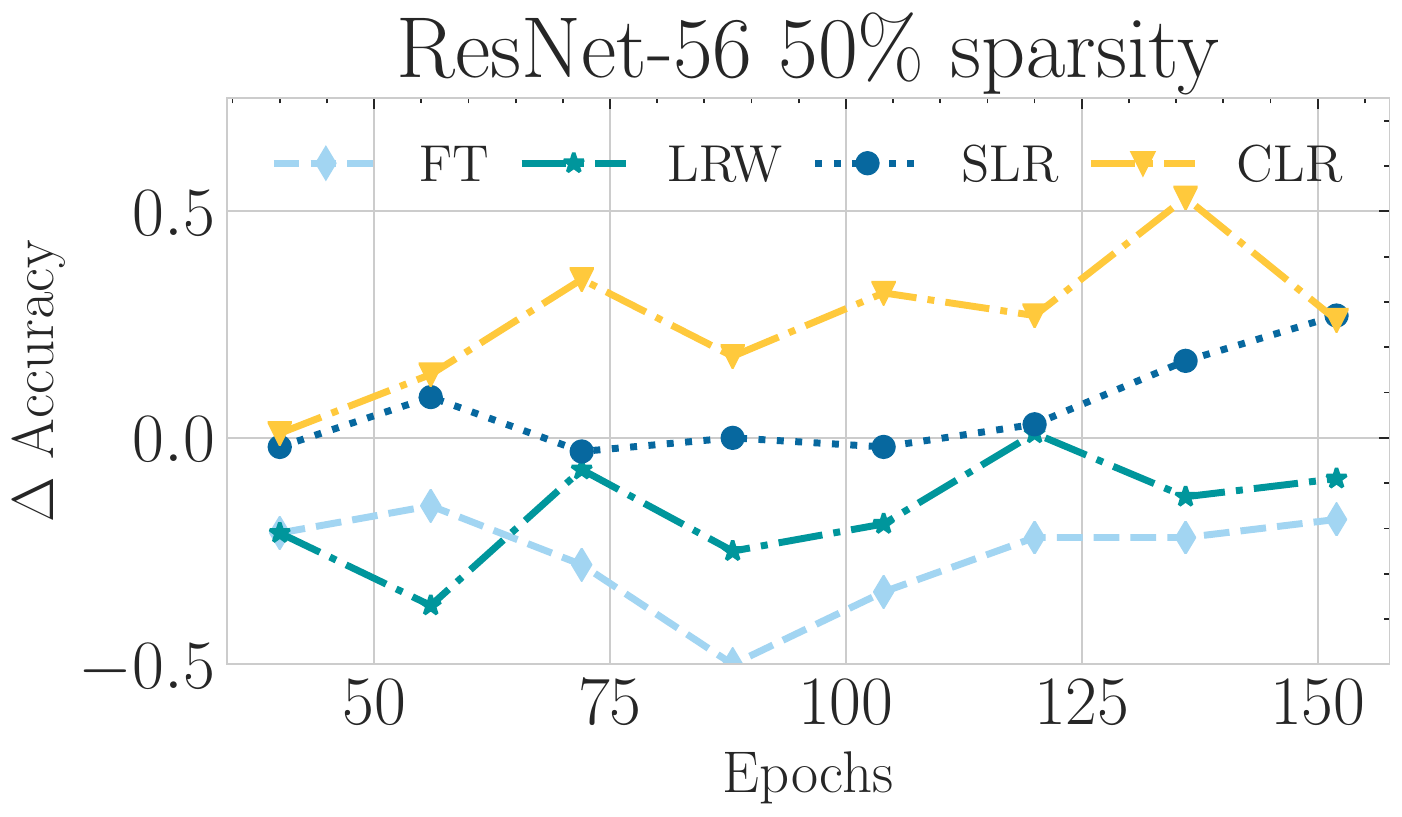}
			\caption{}
			\label{fig:oneshot_mwp_cifar10_a}
		\end{subfigure}
		\hfill
		\begin{subfigure}[b]{\sc\textwidth}
			\centering
			\includegraphics[width=\textwidth]{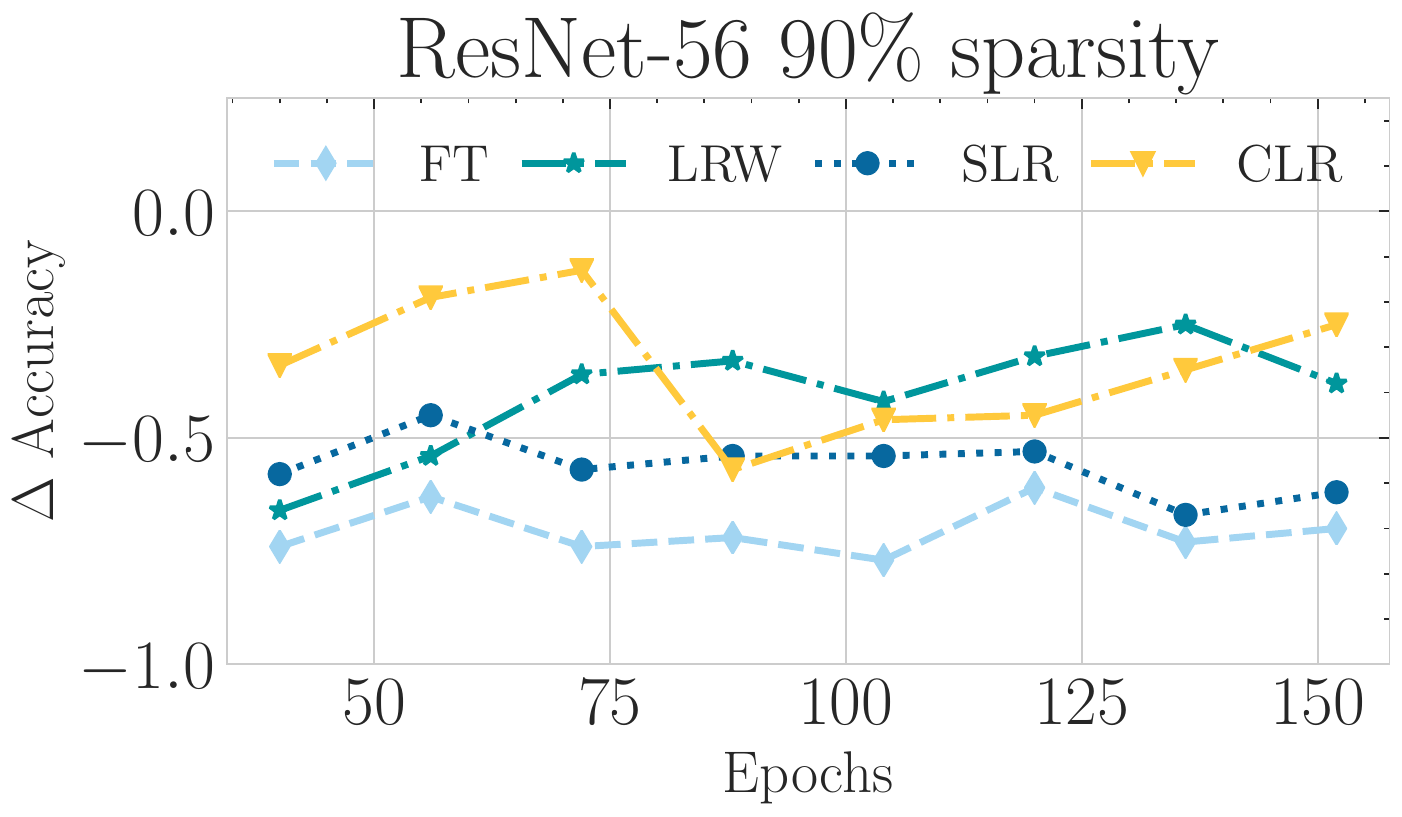}
			\caption{}
			\label{fig:oneshot_mwp_cifar10_b}
		\end{subfigure}
		\hfill
		\begin{subfigure}[b]{\sc\textwidth}
			\centering
			\includegraphics[width=\textwidth]{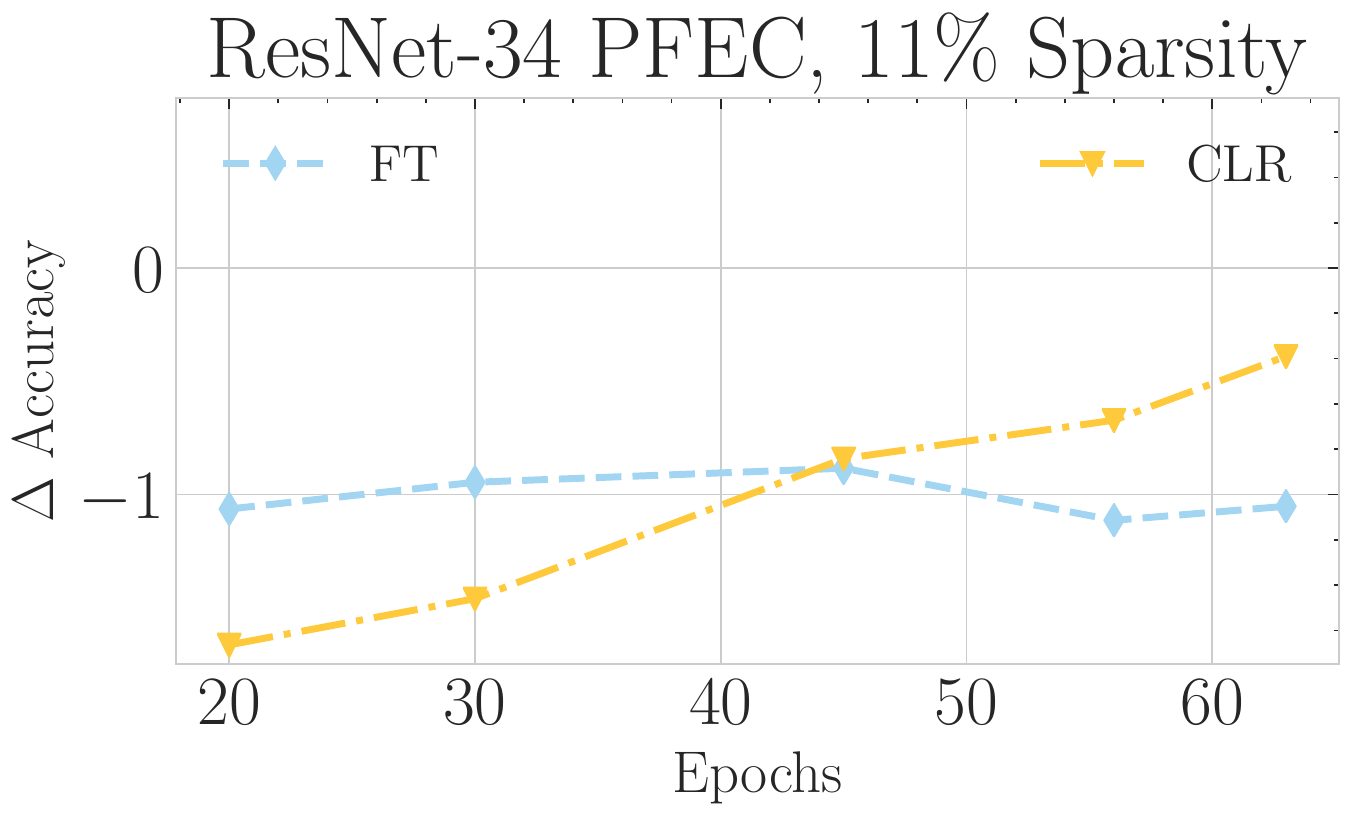}
			\caption{}
			\label{fig:oneshot_mwp_cifar10_b}
		\end{subfigure}
		
		\caption{\textit{One-shot} \textit{unstructured} pruning on CIFAR-10 dataset using MWP \citep{han2015learning} ((a) and (b)) and \textit{structured} pruning on ImageNet with $\ell_1$-norm filters pruning \citep{li2016pruning} (c). }
		\label{fig:oneshot_mwp_cifar10}
	\end{figure}
	
	\subsection{Retraining cost and Performance Trade-off}
	We first investigate the performance of retraining techniques while varying the retraining budget. 
	Figure~\ref{fig:oneshot_pfec_cifar10_original} illustrates the results of $\ell_1$-norm filters pruning (PFEC)~\citep{li2016pruning} with different retraining techniques on CIFAR-10. We can observe that, larger values of learning rate always attain higher performance than \textit{fine-tuning} regardless of number of retraining epochs. Furthermore, with some additional retraining budgets, e.g., 80 to 120 epochs compare to 40 epochs in original works, we can attain much higher compression ratio with almost no accuracy drop (see Figure~\ref{fig:oneshot_pfec_cifar10_compression} in Appendix). 
	
	Figure~\ref{fig:oneshot_mwp_cifar10} reports the accuracy of ResNet-56 pruned with magnitude-based weights pruning~\citep{han2015learning} in a one-shot manner. We study the impact of different retraining schedules on low and high sparsity i.e. $50\%$ and $90\%$ respectively. It can be observed that under both setting learning rate restarting consistently outperforms fine-tuning across number of retraining epochs.
	
	To verify the effectiveness of CLR on large-scale datasets, we conduct experiment using $\ell_1$-norm filters pruning with ResNet-34 on ImageNet in Figure~\ref{fig:oneshot_mwp_cifar10}(c). We find that CLR achieve lower performance than fine-tuning for low epochs. This is expected phenomenon since \citep{huang2017snapshot} also found that learning rate restarting on ImageNet usually requires training for $45$ epochs while it only needs $40$ epochs on CIFAR-10. When retraining with number of epochs higher than $45$, CLR can reach higher accuracy than fine-tuning.
	
	Thus, the value of LRW can be a good heuristic to choose the learning rate for retraining after pruning. A concrete example is that we achieve significant gains with Taylor pruning when retraining for only $25$ epochs as shown in Table \ref{table:taylor-pruning}.
	
	\subsection{Model Size and Performance Trade-off}
	In this section, we investigate the trade-off of between model size and performance of compact models under a fixed retraining budgets. \cite{renda2020comparing} found that learning rate rewinding usually saturate at half of original training, thus, we perform on retraining for $80$ epochs on CIFAR-10 and $45$ epochs on ImageNet. 
	
\begin{figure}
		\centering
		\begin{subfigure}[b]{0.325\textwidth}
			\centering
			\includegraphics[width=\textwidth]{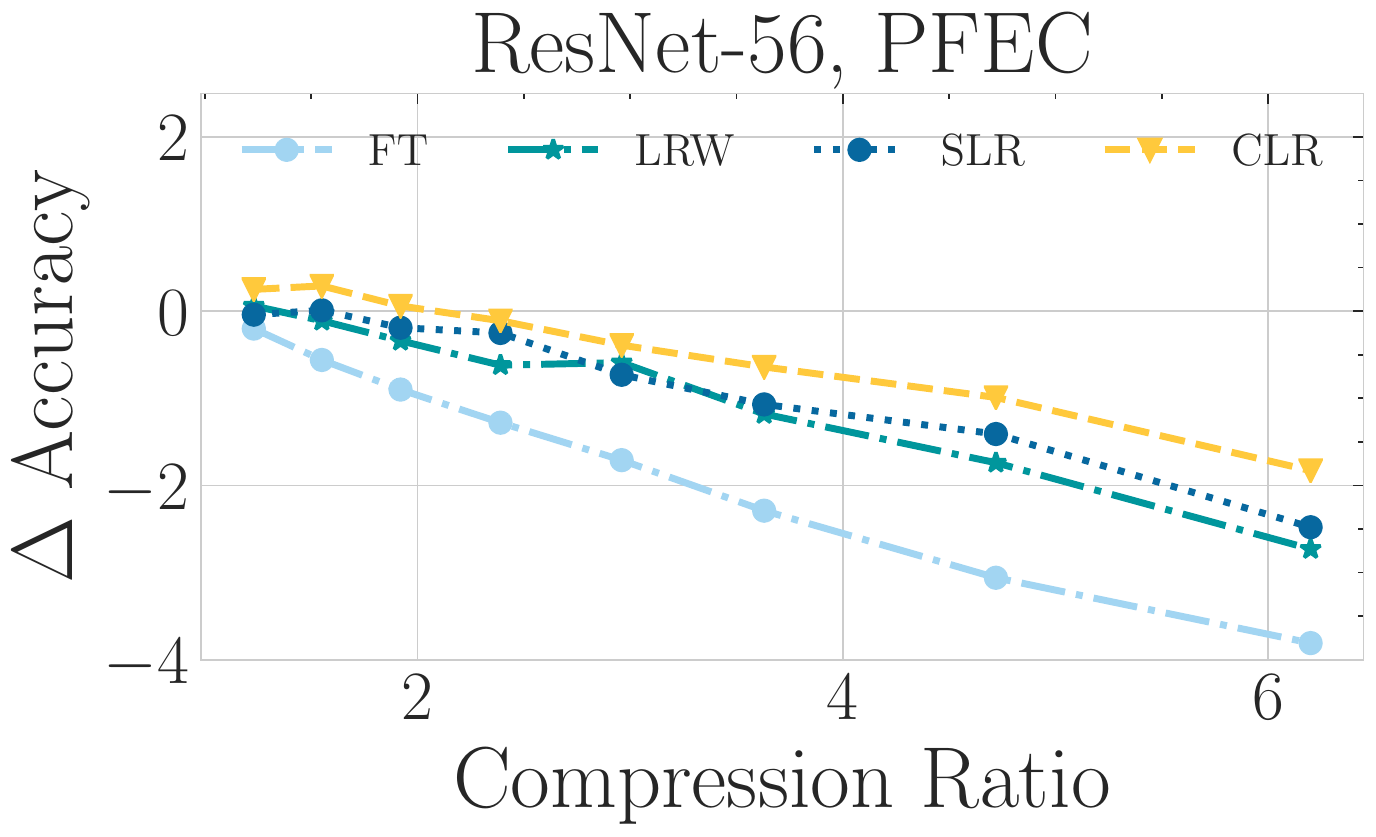}
			\caption{}
			{\tiny {\tiny }}			\label{fig:oneshot_mwp_cifar10_a}
		\end{subfigure}
		\hfill
		\begin{subfigure}[b]{0.325\textwidth}
			\centering
			\includegraphics[width=\textwidth]{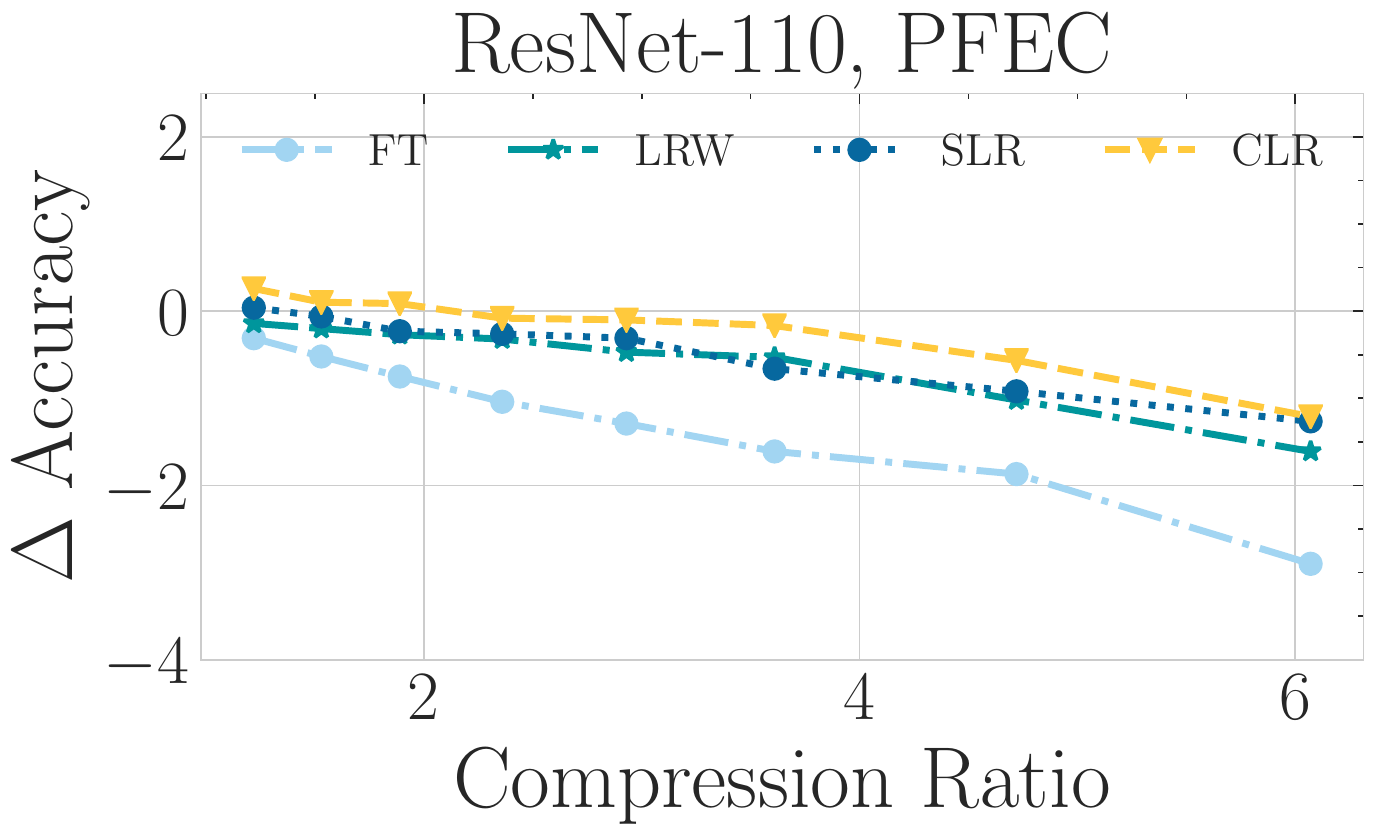}
			\caption{}
			\label{fig:oneshot_mwp_cifar10_b}
		\end{subfigure}
		\hfill
		\begin{subfigure}[b]{0.325\textwidth}
			\centering
			\includegraphics[width=\textwidth]{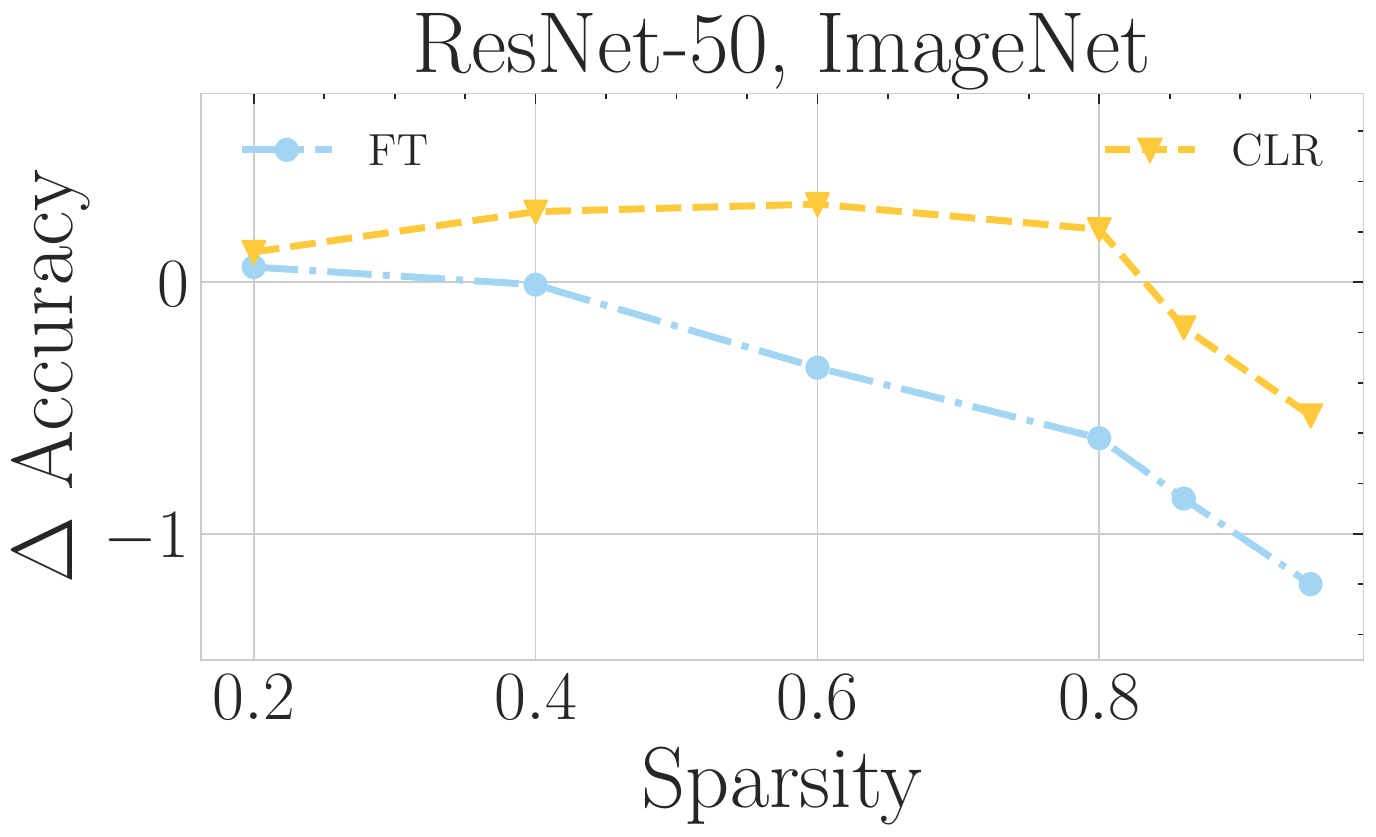}
			\caption{}
			\label{fig:oneshot_mwp_cifar10_c}
		\end{subfigure}
		\caption{\textit{Iterative} pruning on CIFAR-10 dataset using $\ell_1$-norm filters pruning~\cite{li2016pruning} ((a) and (b)) and on ImageNet using magnitude-based weights pruning \citep{han2015learning} ((c)).}
		\label{fig:iterative_pfec_cifar10}
	\end{figure}

	We first experiment with iterative $\ell_1$-norm filters pruning on CIFAR-10 and report the results in Figure \ref{fig:iterative_pfec_cifar10}(a,b), we can observe that {SLR} and {CLR} also perform comparable or better than {LRW} in this setting. We then conduct experiments with iterative unstructured pruning on ImageNet with ResNet-50 to verify the superior of CLR compared to fine-tuning in Figure \ref{fig:iterative_pfec_cifar10}(c), we find that MWP with CLR significantly outperform fine-tuning and even increase the accuracy of the network until $80\%$ sparsity (of convolutional layers).
	
	In our experiments, CLR usually reaches slightly higher accuracy than LRW and SLR. A possible explanation is that 1-cycle schedule gives better convergence speed than conventional schedules  \citep{smith2019super}, thus, bringing much better results when employing ``low" retraining budgets. 
	
	\subsection{Evaluation with Other Pruning Algorithms}
	\label{appendix:sophisticated_pruning_clr}
	We investigate whether the effectiveness of learning rate restarting holds with other pruning algorithms than simple norm-based pruning. As the performance of larger learning rate schedules such as LRW, SLR, and CLR are rather similar, we select CLR for use in this experiment. Particularly, we examine \textit{CLR} with SFP \citep{he2018soft} in this section..
	
	
	\textbf{Soft Filters Pruning} \citep{he2018soft}.
	Previous sections demonstrate that the success of large learning rate hold for a wide range of pruning algorithms, network architectures and datasets. However, we note that all examined pruning algorithms followed the \emph{pruning-retraining} paradigm, i.e., it is necessary to retrain the pruned model.
	In this part, we investigate \emph{regularizing} where we gradually, softly prune the network during the course of training. In contrast to \textit{pruning-retraining}, the latter approach does not require further training after final pruning. That being said, we still make a case study to verify if retraining does improve the performance of models pruned with this type of algorithms.
	
	To make a fair comparison between fine-tuning and no fine-tuning, we randomly split the conventional training set of CIFAR-10/CIFAR-100 (including $50000$ images) to \emph{train} ($90\%$ images of total set) and \emph{val} ($10\%$ remaining images) set and we then report the result of best-validation models on ``standard" \emph{test} set (including $5000$ images).
	
	\begin{table}[t]
		\caption{Results (accuracy) for soft filters pruning \citep{he2018soft}.  The ``w/o FT'' column demonstrate the performance of networks follow original work (do not retrain after final prune). ``FT'' column indicates results of networks when fine-tuning for $200$ more epochs. ``CLR-x'' columns show accuracy of networks after retraining with CLR for $x$ more epochs.``$\%$PF" indicates number of pruned filters.
		}. 
		\small
		\centering
		\begin{tabular}{l | c lll l l}
			\toprule
			Model    &   $\%$PF  &  w/o FT &  FT &  CLR-50  & CLR-100 & CLR-200\\ \midrule
			
			\multirow{2}{*}{\shortstack{CIFAR-10\\ResNet-56}}  & $30$ & 92.74$\pm0.28$   & 92.85$\pm0.21$ & 92.57$\pm0.19$ & 93.00$\pm0.02$ & $\mathbf{93.18\pm0.25}$ \\
			&  $40$   & 91.95$\pm0.29$ &  91.94$\pm0.26$ &  92.19$\pm0.16$ & 92.23$\pm0.12$ & $\mathbf{92.78\pm0.34}$ \\
			
			\midrule
			
			\multirow{2}{*}{\shortstack{CIFAR-10\\ResNet-110}} &  &  & & &\\
			& 40 & 92.25$\pm0.55$ & 92.23$\pm0.50$ & 92.30$\pm0.37$ & 92.79$\pm0.38$ & $\mathbf{92.91\pm0.41}$\\ \midrule
			
			\multirow{2}{*}{\shortstack{CIFAR-100\\ResNet-56}}  & $30$ &  68.92 $\pm0.33$   & 69.41$\pm0.38$ &70.27$\pm0.40$ & 70.24$\pm0.43$ & $\mathbf{70.29\pm0.46}$ \\
			&  $40$    &  67.69$\pm0.41$ &  67.64$\pm0.50$ &  68.78$\pm0.23$ & $\mathbf{69.12\pm0.22}$ & 68.88$\pm0.44$ \\
			
			\midrule 
			
			\multirow{2}{*}{\shortstack{CIFAR-100\\ResNet-110}} & $30$ &  70.93$\pm0.58$  & 70.94$\pm0.58$ & 71.77$\pm0.60$ & 71.74$\pm0.63$ & $\mathbf{72.00\pm0.21}$\\
			& 40 & 68.41$\pm0.26$ & 69.04$\pm0.41$ & 70.22$\pm0.14$ & 70.05$\pm0.27$ & $\mathbf{70.34\pm0.24}$\\
			\bottomrule
		\end{tabular}
		\label{table:soft-pruning}
	\end{table}
	
	We report the performance of pruned network  without fine-tuning, with fine-tuning and with learning rate restarting (CLR) in Table \ref{table:soft-pruning}. We find that applying fine-tuning on these model usually give very fast convergences speed (roughly in first 20 epochs). However, fine-tuned models exhibit negligible gain then pruned models without fine-tuning. In contrast, CLR significantly improves the performance even with small retraining budgets.
	Interested readers can find our results with other pruning algorithms namely HRank \citep{lin2020hrank} and Taylor pruning \citep{molchanov2019importance} in the Appendix~\ref{appendix:other_pruning}.
	
	\subsection{Scaling Original Training}
	\cite{liu2018rethinking} present empirical evidence that 3-stage pipeline of \textit{structured} pruning is inferior to training compact models from scratch if training budgets of compact models are scaled proportion to FLOPs ratio between original and pruned model. Following \cite{evci2019rigging, gale2019state}, we conduct experiments to compare this strong baseline with $\ell_1$-norm filters pruning while employing CLR on CIFAR-10, CIFAR-100 and ImageNet. Specifically, we compare pruned networks retrained with fine-tuning and CLR mechanism with Scratch-B(udget) and Scratch-E(pochs) which train pruned networks from scratch with the same computational budget and epochs respectively \footnote{For Scratch-E, we use $160$ (original training) + $40$ (retraining) = $200$ epochs while training from scratch instead of $160$ as in \citep{liu2018rethinking}}.
	
	When evaluating with CIFAR-10 and CIFAR-100, we found that the validation accuracies of Scratch-B, Scratch-E, and CLR are very similar to each other, and therefore, the comparisons and ranking could be very sensitive to the variance of the accuracy. To reduce variance, we employ the following training scheme. First, we re-run each experiment $5$ times on CIFAR-10 and CIFAR-100 (more than previous experiments to reduce noise). Second, we randomly (and independently between each run) split the original training set of CIFAR-10 to \textit{training} (90\% images of total set) and \textit{validation} set (10\% images of total set) and report the performance of \textit{best validation} models on original \textit{test set}.
	Note that for ImageNet, we re-run the experiments $3$ times. We found that for ImageNet, the variance is negligible and thus we simply report the \textit{best validation} accuracy.
	
	
	For retraining, we adopt the same retraining budgets as in \citep{liu2018rethinking, li2016pruning} ($40$ and $20$ epochs for CIFAR-10 and ImageNet respectively). Since the total number of retraining epochs is limited, for ImageNet, we restart the learning rate to $0.01$ (instead of $0.1$ in the previous section) to guarantee the convergence of the models.
	
	The results of pruned network via PFEC are shown in Table \ref{table:rethinking_pfec}. We can observe that for CIFAR-10, the results of fine-tuning and Scratch-B/E are aligned with the prior work of \cite{liu2018rethinking} while CLR achieve slightly better results than Scratch-E and comparable with Scratch-B. 
	However, PFEC+CLR significantly outperform both fine-tuning and Scratch-B by a large margin.
	\begin{table}[t]
		\small
		\caption{Results (accuracy) for $\ell_1$-norm based filter pruning \citep{li2016pruning}. Configurations of Model and Pruned Model are both from the original paper \citep{li2016pruning}. The results of  ``Scratch-E" and ``Scratch-B" on ImageNet are taken directly from work of \cite{liu2018rethinking}. Top- and second-ranked results are highlighted in \textbf{\textcolor{blue}{bold blue}} and \textcolor{blue}{blue}.
		}
		\centering
		\resizebox{\columnwidth}{!}{%
			\begin{tabular}{c|c|cccccc}
				\toprule
				\textbf{Dataset} & \textbf{Model}         &  \textbf{Unpruned}    & \textbf{Version}    &  \textbf{Fine-tuning} & \textbf{Scratch-E} & \textbf{Scratch-B} & \textbf{CLR} \\ \midrule
				\multirow{5}{*}{CIFAR-10} & VGG-16 & $93.01\pm0.21$ & -  & $93.09\pm0.19$ & $93.01\pm0.27$ & $\best{93.24\pm0.25}$ & $\secondbest{93.12\pm0.26}$\\
				\cmidrule{2-8}
				& \multirow{2}{*}{ResNet-56} & \multirow{2}{*}{$92.32\pm0.32$}  & A & $92.18\pm0.28$ & $92.60\pm0.17$ & $\best{92.77\pm0.19}$  & $\secondbest{92.71\pm0.33}$ \\
				&   &  & B & $92.06\pm0.23$ & $\secondbest{92.39\pm0.22}$ & $92.31\pm0.23$ & $\best{92.41\pm0.19}$ \\
				\cmidrule{2-8}
				&  \multirow{2}{*}{ResNet-110} & \multirow{2}{*}{$92.91\pm0.38$} & A & $92.81\pm0.39$  & $92.97\pm0.22$ & $\secondbest{93.04\pm0.18}$ & $\best{93.08\pm0.38}$\\
				&   &   & B & $92.34\pm0.33$& $92.63\pm0.70$ & $\best{93.20\pm0.25}$ & $\secondbest{93.03\pm0.31}$\\
				\midrule
				\multirow{5}{*}{CIFAR-100} & VGG-16 & $71.26\pm0.23$& - & $69.97\pm0.32$ & $\best{71.37\pm0.07}$ & $\secondbest{71.72\pm0.34}$ & $71.18\pm0.07$ \\
				\cmidrule{2-8}
				& \multirow{2}{*}{ResNet-56} & \multirow{2}{*}{$69.51\pm0.25$} & A & $69.55\pm0.54$ & $\best{70.15\pm0.52}$ & $69.74\pm0.57$& $\secondbest{70.04\pm0.30}$\\
				&   &  & B & $69.53\pm0.28$ & $69.64\pm0.24$ & $\secondbest{69.91\pm0.53}$ & $\best{70.06\pm0.35}$ \\
				\cmidrule{2-8}
				&  \multirow{2}{*}{ResNet-110} & \multirow{2}{*}{$70.59\pm0.36$} & A & $70.71\pm0.16$ & $\secondbest{71.11\pm0.05}$ & $70.61\pm0.25$ & $\best{71.29\pm0.25}$ \\
				&   &   & B & $69.54\pm0.27$ & $\secondbest{70.61\pm0.45}$ & $\best{70.79\pm0.22}$ & $70.52\pm0.31$ \\
				\midrule
				\multirow{2}{*}{ImageNet}  & \multirow{2}{*}{ResNet-34} &  \multirow{2}{*}{$73.30$} & A &  $72.96\pm0.06$  &  $72.56$ & $\secondbest{73.03}$ &  $\best{73.44\pm0.06}$  \\
				&  &  & B &  $72.50\pm0.08$  & $72.29$ & $\secondbest{72.91}$ &  $\best{73.14\pm0.03}$ \\
				
				\bottomrule
			\end{tabular}
		}
		\label{table:rethinking_pfec}
	\end{table}
	
	\textbf{Discussion.} Our empirical results suggest that practitioners should employ retraining with large learning rate schedules as an alternative technique for fine-tuning to obtain compact models with better performance after network pruning. In our results, cyclic learning rate restarting (CLR) is slightly more efficient than scaled learning rate restarting (SLR) and learning rate rewinding (LRW).

	\section{Interplay of Pruning Algorithms and Retraining Schemes}
	\label{sec:pitfalls}
	Section \ref{sec:case_study} suggests that learning rate schedule can significantly improve the performance of several pruning algorithms. We notice that there are notable differences between settings of implementation of network pruning especially in retraining phase. 
	In this section, we show that the difference between implementation could easily lead to misleading results and unfair comparisons between pruning algorithms.
	
	\subsection{A Strong Baseline: $\ell_1$-norm Filters Pruning with CLR}
	\label{sec:strong_baseline_l1}
	In this section, we demonstrate that even with \textbf{same} retraining budgets, utilizing simple \textit{CLR} with $\ell_1$-norm filters pruning can achive comparable or exceed the performance of more sophisticated saliency metrics \emph{without} meticulous hyperparameters searching. The implementation details are as follows.
	\paragraph{Our baseline} For pruning, we adopt the pretrained of Torchvision and apply $\ell_1$-norm Filters Pruning on these models. In our implementation, the number of removed filters in each block are approximately equal so that the final pruned models have similar number of parameters with the compared one. Thus, if any, this setting should favor other approaches involving laborious sensitive analysis.
	For retraining, we apply the \textit{CLR} learning rate schedule while choosing the maximum value of learning rate according to \emph{learning rate rewinding}. For pruning algorithms that only retrain for small number of epochs (e.g. $25$ epochs in case of Taylor Pruning) we also \textit{restart} the learning rate value to the slightly higher value of $0.01$.
	
	\paragraph{Generative Adversarial Learning (GAL)} \citet{lin2019towards} suggest to leverage generative adversarial learning, which learns a sparse soft mask in a label-free and an end-to-end manner, to effectively solve the optimization problem. The original work retrain the pruned network for $30$ epochs with batch size of $32$ and initial learning rate of $0.01$ (which is decayed by factor of $10$ every $10$ epochs). Thus, we use \textit{CLR} with learning rate of $0.01$ and batch size of $32$ in our experiments when comparing with \textit{GAL}. Table \ref{table:gal_pfec} shows the results of PFEC + CLR when retrained under same settings with GAL. We can see that PFEC+CLR can create a very strong baseline compare with GAL. 
	
	\begin{table}[!t]
		\caption{Comparing the performance of pruned network via PFEC + CLR and GAL on ImageNet. The results of GAL are taken directly from original papers.
		}
		\small 
		\centering
		\begin{tabular}{c|c|c|cc|c}
			\toprule
			\textbf{Model}         &  \textbf{Unpruned Top-1}    & \textbf{Method}    &  \textbf{Param} $\downarrow\%$ & \textbf{FLOPs} $\downarrow\%$             &    \textbf{Top-1} \\ \midrule
			\multirow{4}{*}{ResNet-50} &   \multirow{4}{*}{$76.15$} & GAL-0.5 & $17.2$ & $43.0$  &  $71.95$ \\
			&  & PFEC + CLR  & $17.6$ & -  & $\mathbf{75.26}$\\
			\cmidrule{3-6}
			&   &  GAL-1 & $42.6$ & $61.4$  &  $69.88$ \\
			&   & PFEC + CLR  & $43.1$ & -  & $\mathbf{73.81}$\\
			
			\bottomrule
		\end{tabular}
		\label{table:gal_pfec}
	\end{table}
	
	\paragraph{HRankPlus} An extension of HRank \citep{lin2020hrank} which is better than HRank in both retraining efficiency and performance \footnote{\url{https://github.com/lmbxmu/HRankPlus} \label{footnote:hrankplus} }.  Inspired by the discovery that average rank of feature maps of a filter is consistent, the authors suggested to iteratively remove low-rank feature maps that contain less information. In the original implementation, the authors retrain the pruned networks for $90$ epochs with initial learning rate of $0.1$  and also utilize label smoothing. For simplicity, we do not use label smoothing in our implementation. As shown in Table \ref{table:hrankplus_pfec}, PFEC+CLR also attain similar results with HRankPlus approach.
	\begin{table}[!t]
		\caption{Comparing the performance of pruned network via PFEC + CLR and HRankPlus on ImageNet. The results of HRankPlus are taken directly from official Github repository. 
		}
		\small 
		\centering
		\begin{tabular}{c|c|c|cc|c}
			\toprule
			\textbf{Model}         &  \textbf{Unpruned Top-1}    & \textbf{Method}    &  \textbf{Param} $\downarrow\%$ & \textbf{FLOPs} $\downarrow\%$             &    \textbf{Top-1} \\ \midrule
			\multirow{4}{*}{ResNet-50} &   \multirow{4}{*}{$76.15$} & HRankPlus & $ 40.8$ & $44.8$  &  $75.56$ \\
			&  & PFEC + CLR  & $41.4$ & -  & $\mathbf{75.59}$\\
			\cmidrule{3-6}
			&   &  HRankPlus & $56.7$ & $62.8$  &  $74.19$ \\
			&   & PFEC + CLR  & $56.9$ & -  & $\mathbf{74.39}$\\
			
			\bottomrule
		\end{tabular}
		\label{table:hrankplus_pfec}
	\end{table}
	
	Additional results on CIFAR-10 and Discrimination-aware Channel Pruning \citep{zhuang2018discrimination}, Taylor Pruning \citep{molchanov2019importance}, Provable Filters Pruning \citep{Liebenwein2020Provable} are reported in Appendix \ref{appendix:strong_baseline_with_pfec} for interested readers.
	
	
	
	\subsection{Random Pruning with Learning Rate Restarting}
	\label{sec:random_pruning}
	
	In last subsection, we demonstrated that slight modification in retraining of simple $\ell_1$-norm Filters Pruning can exceed the perfomance of much more sophisticated pruning algorithms. In this subsection, we investigate the interplay between pruning saliency metrics and retraining configurations by comparing accuracy of randomly pruned networks with the \emph{original performance} of methodically pruned networks.
	
	We directly adopt the original implementation from the authors of Taylor Pruning, HRankPlus for comparisons. For PFEC and MWP we make use of the implementation of \cite{liu2018rethinking}. 
	We selected these works because they do not utilize a similar learning rate value as LRW in the original implementation, i.e., the authors applied a smaller value then the heuristic of LRW. The detailed training recipe of the original works can be found in Appendix \ref{appendix:original_random_pruning}.
	To make the randomly pruned network have the same structure as the methodically pruned networks, i.e., equal number of filters per layer, we propose to replace the importance score of each neuron estimated by these methods by a uniformly random score value.

	For retraining randomly pruned networks, we adopt the same principle at described in Section \ref{sec:strong_baseline_l1}. Table \ref{table:random_imagenet} presents novel results of performance of random pruning + CLR with corresponding results of considered pruning algorithms. It is worth pointing out that we report the \textit{original} results of all methodical pruning are taken directly from their papers respectively. In our experiments, Random Pruning + CLR could surpass the performance of sophisticated saliency metrics even on large-scale and challenging dataset such as ImageNet with only minimal change in learning rate schedule and initial learning rate value.
	
	Comprehensive results of random pruning with different compression ratio for $\ell_1$-norm Filters Pruning and MWP is reported in Appendix \ref{appendix:random_pruning_compression_ratio}.
	
	\begin{table}[t]
		\caption{Results of networks when applying random pruning and methodically pruning algorithms. ``Original" column presents accuracy of pruned network reported in original papers. ``R-CLR" presents the results of Random Pruning with CLR.
		}
		\small
		\centering
		\resizebox{\columnwidth}{!}{%
			\begin{tabular}{c|c|l|c c cc} 
				\toprule
				\textbf{Dataset} & \textbf{Method} & \textbf{Model}    & \textbf{Params} $\downarrow$ \% &  \textbf{FLOPs} $\downarrow \%$  &  \textbf{Original} &  \textbf{R-CLR} \\ \midrule
				
				\multirow{4}{*}{CIFAR-10} & \multirow{4}{*}{\shortstack{HRankPlus}} & ResNet-56  & $70.0$ &  $74.1$  &  $92.32$  & $\mathbf{92.40\pm0.16}$ \\
				& & ResNet-110  & $68.3$ &  $71.6$ & $93.23$ & $\mathbf{93.37\pm0.04}$  \\
				& & DenseNet-40  & $61.9$ &  $59.9$ & $93.66$   & $\mathbf{93.71\pm0.05}$ \\
				& & VGG-16  & $87.3$ &  $78.6$ &  $\mathbf{93.10}$  &  $93.06\pm0.07$ \\ 
				
				\midrule
				
				\multirow{6}{*}{ImageNet} & \multirow{2}{*}{PFEC}  &  ResNet-34 A & $2.3$ & $15.9$  &  $72.96\pm0.06$  & $\mathbf{73.47\pm0.08}$ \\
				& & ResNet-34 B & $10.8$ & $24.2$  &  $72.50\pm0.08$   & $\mathbf{73.05\pm0.05}$ \\
				\cmidrule{2-7}
				& \multirow{3}{*}{\shortstack{Taylor \\ Pruning}} & ResNet-50 $72\%$ & $44.5$ & $45.0$ & $74.50$ & $\mathbf{74.91}$  \\
				& & ResNet-50 $81\%$ & $30.1$ & $35.0$ & $75.48$ &  $\mathbf{75.54}$ \\
				& & ResNet-50 $91\%$ & $11.4$ & $20.0$ & $\mathbf{76.43}$ &  $75.93$ \\
				\bottomrule
				
			\end{tabular}
		}
		\label{table:random_imagenet}
	\end{table}
	
	
	These results suggest that retraining techniques, e.g., learning rate restarting and learning rate schedule, play a pivotal role to final performance. Thus, in order to perform fair comparison of different methods, one should be cautious of this seemingly subtle detail. While it is unclear to set up a fair comparison between pruning algorithms that belong to different categories, e.g., pruning before/during/after training, iterative vs. oneshot pruning, we advocate standardizing (a set of) retraining configurations for each algorithm group or thorough hyperparamters searching to find the best configuration of each pruning method when evaluating different algorithms.
	
	\section{Discussion and Conclusion}
	In this work, we conducted extensive experiments to show that learning rates do matter in achieving good performance of pruned neural networks. We concluded that compared to traditional fine-tuning, learning rate restarting in general is an efficient way to retrain pruned networks to recover performance drop due to pruning. The success of learning rate rewinding is accounted by the use of large learning rates in the retraining. We believe that these findings help raise awareness of proper use of learning rate schedule when desiging pruning algorithms, standardizing empirical experiments and allowing fair comparisons. Our takeaway message is:
	\begin{center}	
		\sl Pruning algorithms should be compared in the same retraining configurations.
	\end{center}
	
	Our work is not without limitations. So far we investigated with hyperparameters identical to those of the original implementations of the pruning algorithms, and only experiment with different learning rate schedules. We also limited our experiments to the image classification task in computer vision. Considering more datasets and other domains is a great research avenue for future work. 
	
	\bibliography{iclr2021_conference}

\begin{thebibliography}{37}
\providecommand{\natexlab}[1]{#1}
\providecommand{\url}[1]{\texttt{#1}}
\expandafter\ifx\csname urlstyle\endcsname\relax
  \providecommand{\doi}[1]{doi: #1}\else
  \providecommand{\doi}{doi: \begingroup \urlstyle{rm}\Url}\fi

\bibitem[Arora et~al.(2018)Arora, Cohen, and Hazan]{arora2018optimization}
Sanjeev Arora, Nadav Cohen, and Elad Hazan.
\newblock On the optimization of deep networks: Implicit acceleration by
  overparameterization.
\newblock \emph{arXiv preprint arXiv:1802.06509}, 2018.

\bibitem[Brutzkus \& Globerson(2019)Brutzkus and Globerson]{brutzkus2019larger}
Alon Brutzkus and Amir Globerson.
\newblock Why do larger models generalize better? a theoretical perspective via
  the xor problem.
\newblock In \emph{International Conference on Machine Learning}, pp.\
  822--830. PMLR, 2019.

\bibitem[Evci et~al.(2019)Evci, Gale, Menick, Castro, and
  Elsen]{evci2019rigging}
Utku Evci, Trevor Gale, Jacob Menick, Pablo~Samuel Castro, and Erich Elsen.
\newblock Rigging the lottery: Making all tickets winners.
\newblock \emph{arXiv preprint arXiv:1911.11134}, 2019.

\bibitem[Frankle \& Carbin(2019)Frankle and Carbin]{frankle2018lottery}
Jonathan Frankle and Michael Carbin.
\newblock The lottery ticket hypothesis: Finding sparse, trainable neural
  networks.
\newblock In \emph{International Conference on Learning Representations}, 2019.

\bibitem[Gale et~al.(2019)Gale, Elsen, and Hooker]{gale2019state}
Trevor Gale, Erich Elsen, and Sara Hooker.
\newblock The state of sparsity in deep neural networks.
\newblock \emph{arXiv preprint arXiv:1902.09574}, 2019.

\bibitem[Gao et~al.(2020)Gao, Huang, Pei, and Huang]{gao2020discrete}
Shangqian Gao, Feihu Huang, Jian Pei, and Heng Huang.
\newblock Discrete model compression with resource constraint for deep neural
  networks.
\newblock In \emph{Proceedings of the IEEE/CVF Conference on Computer Vision
  and Pattern Recognition}, pp.\  1899--1908, 2020.

\bibitem[Han et~al.(2015)Han, Pool, Tran, and Dally]{han2015learning}
Song Han, Jeff Pool, John Tran, and William Dally.
\newblock Learning both weights and connections for efficient neural network.
\newblock In \emph{Advances in neural information processing systems}, pp.\
  1135--1143, 2015.

\bibitem[He et~al.(2018)He, Kang, Dong, Fu, and Yang]{he2018soft}
Yang He, Guoliang Kang, Xuanyi Dong, Yanwei Fu, and Yi~Yang.
\newblock Soft filter pruning for accelerating deep convolutional neural
  networks.
\newblock \emph{arXiv preprint arXiv:1808.06866}, 2018.

\bibitem[He et~al.(2019)He, Liu, Wang, Hu, and Yang]{he2019filter}
Yang He, Ping Liu, Ziwei Wang, Zhilan Hu, and Yi~Yang.
\newblock Filter pruning via geometric median for deep convolutional neural
  networks acceleration.
\newblock In \emph{Proceedings of the IEEE Conference on Computer Vision and
  Pattern Recognition}, pp.\  4340--4349, 2019.

\bibitem[He et~al.(2017)He, Zhang, and Sun]{he2017channel}
Yihui He, Xiangyu Zhang, and Jian Sun.
\newblock Channel pruning for accelerating very deep neural networks.
\newblock In \emph{Proceedings of the IEEE International Conference on Computer
  Vision}, pp.\  1389--1397, 2017.

\bibitem[Huang et~al.(2017)Huang, Li, Pleiss, Liu, Hopcroft, and
  Weinberger]{huang2017snapshot}
Gao Huang, Yixuan Li, Geoff Pleiss, Zhuang Liu, John~E Hopcroft, and Kilian~Q
  Weinberger.
\newblock Snapshot ensembles: Train 1, get m for free.
\newblock \emph{arXiv preprint arXiv:1704.00109}, 2017.

\bibitem[Kusupati et~al.(2020)Kusupati, Ramanujan, Somani, Wortsman, Jain,
  Kakade, and Farhadi]{kusupati2020soft}
Aditya Kusupati, Vivek Ramanujan, Raghav Somani, Mitchell Wortsman, Prateek
  Jain, Sham Kakade, and Ali Farhadi.
\newblock Soft threshold weight reparameterization for learnable sparsity.
\newblock \emph{arXiv preprint arXiv:2002.03231}, 2020.

\bibitem[LeCun et~al.(1990)LeCun, Denker, and Solla]{lecun1990optimal}
Yann LeCun, John~S Denker, and Sara~A Solla.
\newblock Optimal brain damage.
\newblock In \emph{Advances in neural information processing systems}, pp.\
  598--605, 1990.

\bibitem[Lee et~al.(2019)Lee, Ajanthan, and Torr]{lee2018snip}
Namhoon Lee, Thalaiyasingam Ajanthan, and Philip Torr.
\newblock Snip: Single-shot network pruning based on connection sensitivity.
\newblock In \emph{International Conference on Learning Representations}, 2019.

\bibitem[Li et~al.(2020)Li, Wu, Su, Wang, and Lin]{li2020eagleeye}
Bailin Li, Bowen Wu, Jiang Su, Guangrun Wang, and Liang Lin.
\newblock Eagleeye: Fast sub-net evaluation for efficient neural network
  pruning.
\newblock \emph{arXiv preprint arXiv:2007.02491}, 2020.

\bibitem[Li et~al.(2016)Li, Kadav, Durdanovic, Samet, and Graf]{li2016pruning}
Hao Li, Asim Kadav, Igor Durdanovic, Hanan Samet, and Hans~Peter Graf.
\newblock Pruning filters for efficient convnets.
\newblock \emph{arXiv preprint arXiv:1608.08710}, 2016.

\bibitem[Liebenwein et~al.(2020)Liebenwein, Baykal, Lang, Feldman, and
  Rus]{Liebenwein2020Provable}
Lucas Liebenwein, Cenk Baykal, Harry Lang, Dan Feldman, and Daniela Rus.
\newblock Provable filter pruning for efficient neural networks.
\newblock In \emph{International Conference on Learning Representations}, 2020.

\bibitem[Lin et~al.(2020{\natexlab{a}})Lin, Ji, Wang, Zhang, Zhang, Tian, and
  Shao]{lin2020hrank}
Mingbao Lin, Rongrong Ji, Yan Wang, Yichen Zhang, Baochang Zhang, Yonghong
  Tian, and Ling Shao.
\newblock Hrank: Filter pruning using high-rank feature map.
\newblock In \emph{Proceedings of the IEEE/CVF Conference on Computer Vision
  and Pattern Recognition}, pp.\  1529--1538, 2020{\natexlab{a}}.

\bibitem[Lin et~al.(2019)Lin, Ji, Yan, Zhang, Cao, Ye, Huang, and
  Doermann]{lin2019towards}
Shaohui Lin, Rongrong Ji, Chenqian Yan, Baochang Zhang, Liujuan Cao, Qixiang
  Ye, Feiyue Huang, and David Doermann.
\newblock Towards optimal structured cnn pruning via generative adversarial
  learning.
\newblock In \emph{Proceedings of the IEEE Conference on Computer Vision and
  Pattern Recognition}, pp.\  2790--2799, 2019.

\bibitem[Lin et~al.(2020{\natexlab{b}})Lin, Stich, Barba, Dmitriev, and
  Jaggi]{Lin2020Dynamic}
Tao Lin, Sebastian~U. Stich, Luis Barba, Daniil Dmitriev, and Martin Jaggi.
\newblock Dynamic model pruning with feedback.
\newblock In \emph{International Conference on Learning Representations},
  2020{\natexlab{b}}.

\bibitem[Liu et~al.(2019)Liu, Sun, Zhou, Huang, and Darrell]{liu2018rethinking}
Zhuang Liu, Mingjie Sun, Tinghui Zhou, Gao Huang, and Trevor Darrell.
\newblock Rethinking the value of network pruning.
\newblock In \emph{International Conference on Learning Representations}, 2019.

\bibitem[Luo et~al.(2017)Luo, Wu, and Lin]{luo2017thinet}
Jian-Hao Luo, Jianxin Wu, and Weiyao Lin.
\newblock Thinet: A filter level pruning method for deep neural network
  compression.
\newblock In \emph{Proceedings of the IEEE international conference on computer
  vision}, pp.\  5058--5066, 2017.

\bibitem[Molchanov et~al.(2016)Molchanov, Tyree, Karras, Aila, and
  Kautz]{molchanov2016pruning}
Pavlo Molchanov, Stephen Tyree, Tero Karras, Timo Aila, and Jan Kautz.
\newblock Pruning convolutional neural networks for resource efficient
  inference.
\newblock \emph{arXiv preprint arXiv:1611.06440}, 2016.

\bibitem[Molchanov et~al.(2019)Molchanov, Mallya, Tyree, Frosio, and
  Kautz]{molchanov2019importance}
Pavlo Molchanov, Arun Mallya, Stephen Tyree, Iuri Frosio, and Jan Kautz.
\newblock Importance estimation for neural network pruning.
\newblock In \emph{Proceedings of the IEEE Conference on Computer Vision and
  Pattern Recognition}, pp.\  11264--11272, 2019.

\bibitem[Neyshabur et~al.(2018)Neyshabur, Li, Bhojanapalli, LeCun, and
  Srebro]{neyshabur2018towards}
Behnam Neyshabur, Zhiyuan Li, Srinadh Bhojanapalli, Yann LeCun, and Nathan
  Srebro.
\newblock Towards understanding the role of over-parametrization in
  generalization of neural networks.
\newblock \emph{arXiv preprint arXiv:1805.12076}, 2018.

\bibitem[Novak et~al.(2018)Novak, Bahri, Abolafia, Pennington, and
  Sohl-Dickstein]{novak2018sensitivity}
Roman Novak, Yasaman Bahri, Daniel~A Abolafia, Jeffrey Pennington, and Jascha
  Sohl-Dickstein.
\newblock Sensitivity and generalization in neural networks: an empirical
  study.
\newblock \emph{arXiv preprint arXiv:1802.08760}, 2018.

\bibitem[Renda et~al.(2020)Renda, Frankle, and Carbin]{renda2020comparing}
Alex Renda, Jonathan Frankle, and Michael Carbin.
\newblock Comparing rewinding and fine-tuning in neural network pruning.
\newblock \emph{arXiv preprint arXiv:2003.02389}, 2020.

\bibitem[Smith \& Topin(2019)Smith and Topin]{smith2019super}
Leslie~N Smith and Nicholay Topin.
\newblock Super-convergence: Very fast training of neural networks using large
  learning rates.
\newblock In \emph{Artificial Intelligence and Machine Learning for
  Multi-Domain Operations Applications}, volume 11006, pp.\  1100612.
  International Society for Optics and Photonics, 2019.

\bibitem[Tanaka et~al.(2020)Tanaka, Kunin, Yamins, and
  Ganguli]{tanaka2020pruning}
Hidenori Tanaka, Daniel Kunin, Daniel~LK Yamins, and Surya Ganguli.
\newblock Pruning neural networks without any data by iteratively conserving
  synaptic flow.
\newblock \emph{arXiv preprint arXiv:2006.05467}, 2020.

\bibitem[Wang et~al.(2020)Wang, Zhang, and Grosse]{Wang2020Picking}
Chaoqi Wang, Guodong Zhang, and Roger Grosse.
\newblock Picking winning tickets before training by preserving gradient flow.
\newblock In \emph{International Conference on Learning Representations}, 2020.

\bibitem[Wen et~al.(2016)Wen, Wu, Wang, Chen, and Li]{wen2016learning}
Wei Wen, Chunpeng Wu, Yandan Wang, Yiran Chen, and Hai Li.
\newblock Learning structured sparsity in deep neural networks.
\newblock In \emph{Advances in neural information processing systems}, pp.\
  2074--2082, 2016.

\bibitem[Wortsman et~al.(2019)Wortsman, Farhadi, and
  Rastegari]{wortsman2019discovering}
Mitchell Wortsman, Ali Farhadi, and Mohammad Rastegari.
\newblock Discovering neural wirings.
\newblock In \emph{Advances in Neural Information Processing Systems}, pp.\
  2684--2694, 2019.

\bibitem[Ye et~al.(2018)Ye, Lu, Lin, and Wang]{ye2018rethinking}
Jianbo Ye, Xin Lu, Zhe Lin, and James~Z Wang.
\newblock Rethinking the smaller-norm-less-informative assumption in channel
  pruning of convolution layers.
\newblock \emph{arXiv preprint arXiv:1802.00124}, 2018.

\bibitem[You et~al.(2019)You, Yan, Ye, Ma, and Wang]{you2019gate}
Zhonghui You, Kun Yan, Jinmian Ye, Meng Ma, and Ping Wang.
\newblock Gate decorator: Global filter pruning method for accelerating deep
  convolutional neural networks.
\newblock In \emph{Advances in Neural Information Processing Systems}, pp.\
  2133--2144, 2019.

\bibitem[Yu et~al.(2018)Yu, Li, Chen, Lai, Morariu, Han, Gao, Lin, and
  Davis]{yu2018nisp}
Ruichi Yu, Ang Li, Chun-Fu Chen, Jui-Hsin Lai, Vlad~I Morariu, Xintong Han,
  Mingfei Gao, Ching-Yung Lin, and Larry~S Davis.
\newblock Nisp: Pruning networks using neuron importance score propagation.
\newblock In \emph{Proceedings of the IEEE Conference on Computer Vision and
  Pattern Recognition}, pp.\  9194--9203, 2018.

\bibitem[Zhu \& Gupta(2017)Zhu and Gupta]{zhu2017prune}
Michael Zhu and Suyog Gupta.
\newblock To prune, or not to prune: exploring the efficacy of pruning for
  model compression.
\newblock \emph{arXiv preprint arXiv:1710.01878}, 2017.

\bibitem[Zhuang et~al.(2018)Zhuang, Tan, Zhuang, Liu, Guo, Wu, Huang, and
  Zhu]{zhuang2018discrimination}
Zhuangwei Zhuang, Mingkui Tan, Bohan Zhuang, Jing Liu, Yong Guo, Qingyao Wu,
  Junzhou Huang, and Jinhui Zhu.
\newblock Discrimination-aware channel pruning for deep neural networks.
\newblock In \emph{Advances in Neural Information Processing Systems}, pp.\
  875--886, 2018.

\end{thebibliography}
	\bibliographystyle{iclr2021_conference}

	\newpage
	\appendix
	\section*{\LARGE{Appendix}}
	\section{Network Pruning Formulation}
	Let us start with a formal definition of the network pruning problem. Let $\boldsymbol\theta \in \mathbb{R}^D$ be the parameters of a (large) neural network that needs to be pruned, $\mathcal{D}$ be the dataset for training. The parameters $\boldsymbol\theta$ is updated to minimize the loss function $\mathcal{L}(\boldsymbol\theta; \mathcal{D})$. Define \textit{sparsity mask} $\mathbf{m} \in \{0, 1\}^D$ indicating if a weight should be kept or removed.  Similar to  \cite{molchanov2016pruning}, we can formulate the pruning algorithms as finding the optimal sparsity mask such that:
	\begin{align}
	\mathbf{m}^{*} = \underset{\mathbf{m}}{\operatorname{argmin}} \hspace{2mm} \mathcal{L}(\boldsymbol\theta^*\odot\mathbf{m}; \mathcal{D}) - \mathcal{L}(\boldsymbol\theta^*; \mathcal{D})
	\end{align}
	where $\odot$ denotes the Hadamard (element-wise) product, $\boldsymbol\theta^*$ represents the (locally) optimal solution of $\boldsymbol\theta$. An alternative approach is minimizing the different between final output of two networks \citep{lin2019towards}:  
	\begin{align}
	\mathbf{m}^{*} = \underset{\mathbf{m}}{\operatorname{argmin}} \hspace{2mm} \norm{ \mathcal{F}(\boldsymbol\theta^*\odot\mathbf{m}; \mathcal{D}) - \mathcal{F}(\boldsymbol\theta^*; \mathcal{D})}^2_2
	\end{align}
	where $\mathcal{F}: \mathbb{R}^D \rightarrow \mathbb{R}^{D'}$ is the function mapping input feature (i.e. images) to the final output (before softmax). 
	
	Theoretically speaking, as the pruning formulation only takes the loss function into account, and there is a caveat that pruning can result in a set of weights that belong to a bad local minima, making the network generalize poorly. In practice, it is common to observe such issues from the performance drop of the pruned model. In such cases, a fine-tuning process can be applied to further optimize the weights, mitigating the accuracy drop. However, as what fine-tuning means, the weights might be only slightly adjusted without any guarantee about the final network performance.
	
	\section{Training Configuration}
	\label{sec:training_configuration}
	The implementation of $\ell_1$-norm filters pruning (PFEC)~\cite{li2016pruning} and magnitude-based weights pruning (MWP)~\cite{han2015learning} are adopted from the work of \cite{liu2018rethinking}. The implementations of other algorithms are taken from the authors' official repositories.

	For simplicity, we adopt Pytorch's pretrained models for ImageNet. The unpruned models (used for $\ell_1$-norm filters pruning) are trained with below configurations. 
	
	
	\begin{table}[h!]
		\caption{Training configuration for unpruned models. 
			To train CIFAR-10, we use Nesterov SGD with $\beta=0.9$, batch size 64, weight decay $0.0001$ for 160 epochs. 
			To train ImageNet, we use Nesterov SGD with $\beta=0.9$, batch size 32,  weight decay $0.0001$ for 90 epochs.
		}
		\small
		\centering
		\begin{tabular}{lll | l | c}  
			\toprule
			\text{Dataset} & \text{Network} & \text{\#Params.} & \multicolumn{1}{c|}{\text{Learning rate (t = training epoch)}} & \text{Test accuracy} \\  \midrule
			\multirow{6}{*}{CIFAR-10} & \multirow{3}{*}{ResNet-56} & \multirow{3}{*}{0.85M}  \multirow{6}{*}{} & \multirow{6}{*}{$\alpha=\begin{cases}0.1 & \text{t} \in [0, 80) \\ 0.01 & \text{t}\in [80, 120) \\ 0.001 & \text{t}\in [120, 160)\end{cases}$} & \multirow{6}{*}{$93.46 \pm 0.21\%$} \\
			&&&&\\
			&&&&\\
			& \multirow{3}{*}{ResNet-110} & \multirow{3}{*}{1.73M} &&\\
			&&&&\\
			&&&&\\ \midrule
			\multirow{6}{*}{ImageNet} & \multirow{2}{*}{ResNet-18} & \multirow{2}{*}{11.69M} & \multirow{6}{*}{$\alpha = \begin{cases} 0.1 & \text{t}\in [0, 30) \\ 0.01 & \text{t}\in [30, 60)\\ 0.001 & \text{t}\in [60, 90)\end{cases}$} & \multirow{2}{*}{$69.76\%$ top-1} \\
			&&&&\\
			& \multirow{2}{*}{ResNet-34} & \multirow{2}{*}{21.8M}    & & \multirow{2}{*}{$73.30\%$ top-1} \\
			&&&&\\ 
			& \multirow{2}{*}{ResNet-50} & \multirow{2}{*}{25.5M}    & & \multirow{2}{*}{$76.15\%$ top-1} \\
			&&&&\\ 
			\bottomrule
		\end{tabular}
	\end{table}	
	
	\subsection{Additional Results}
\begin{figure}[t]
		\centering
		\def\sc{0.325}
		\begin{subfigure}[b]{\sc\textwidth}
			\centering
			\includegraphics[width=\textwidth]{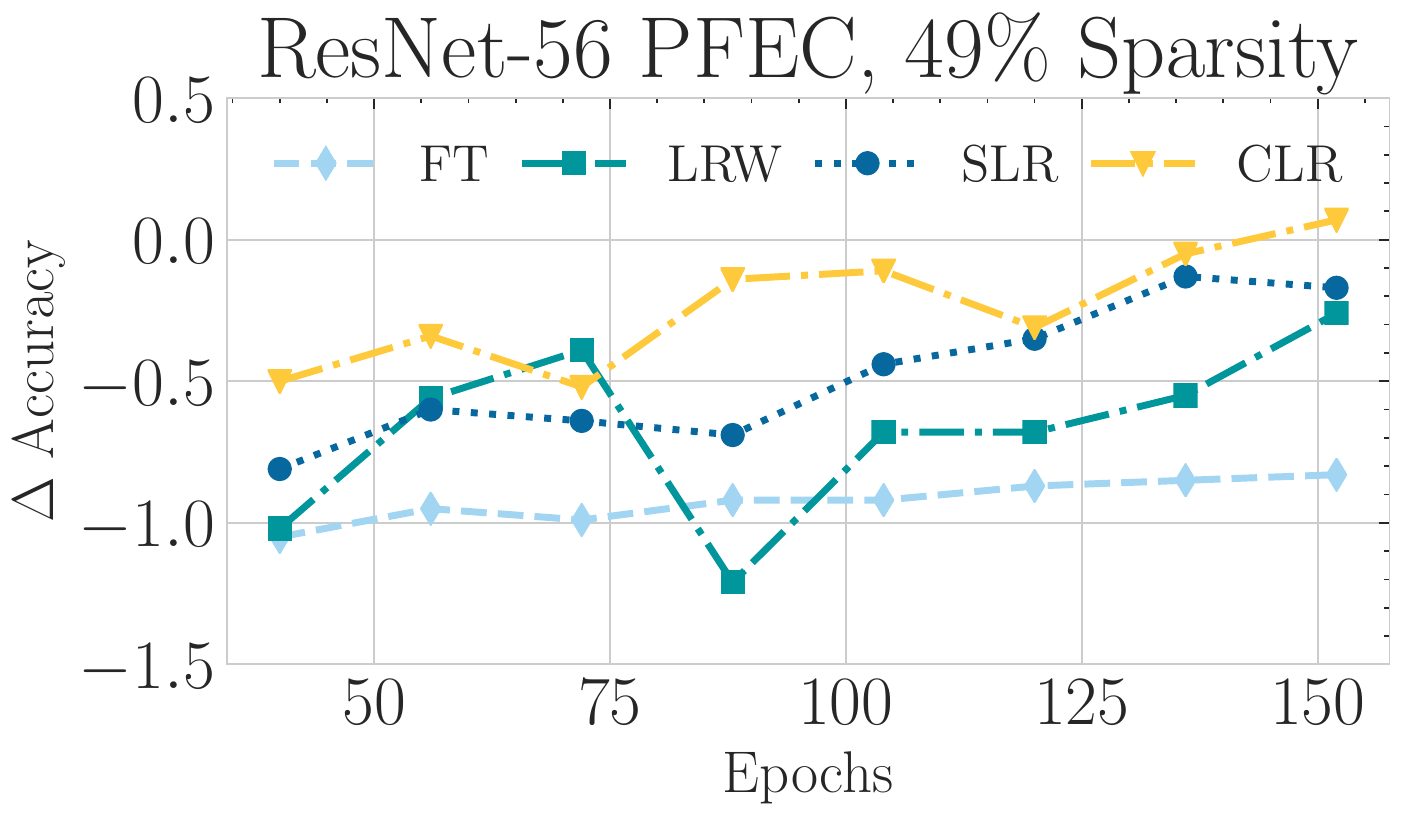}
			\label{fig:oneshot_pfec_cifar10_d}
		\end{subfigure}
		\hfill
		\begin{subfigure}[b]{\sc\textwidth}
			\centering
			\includegraphics[width=\textwidth]{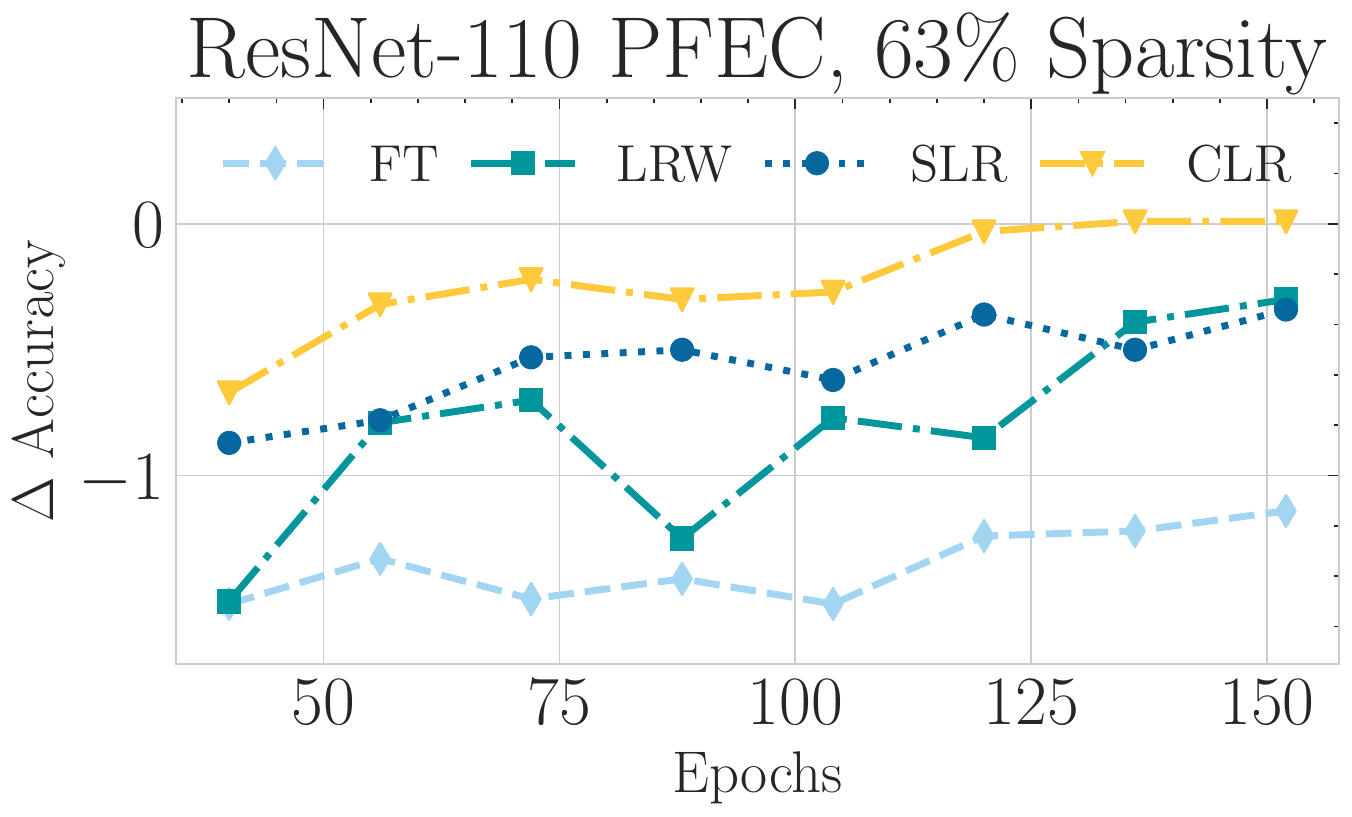}
			\label{fig:oneshot_pfec_cifar10_e}
		\end{subfigure}
		\hfill
		\begin{subfigure}[b]{\sc\textwidth}
			\centering
			\includegraphics[width=\textwidth]{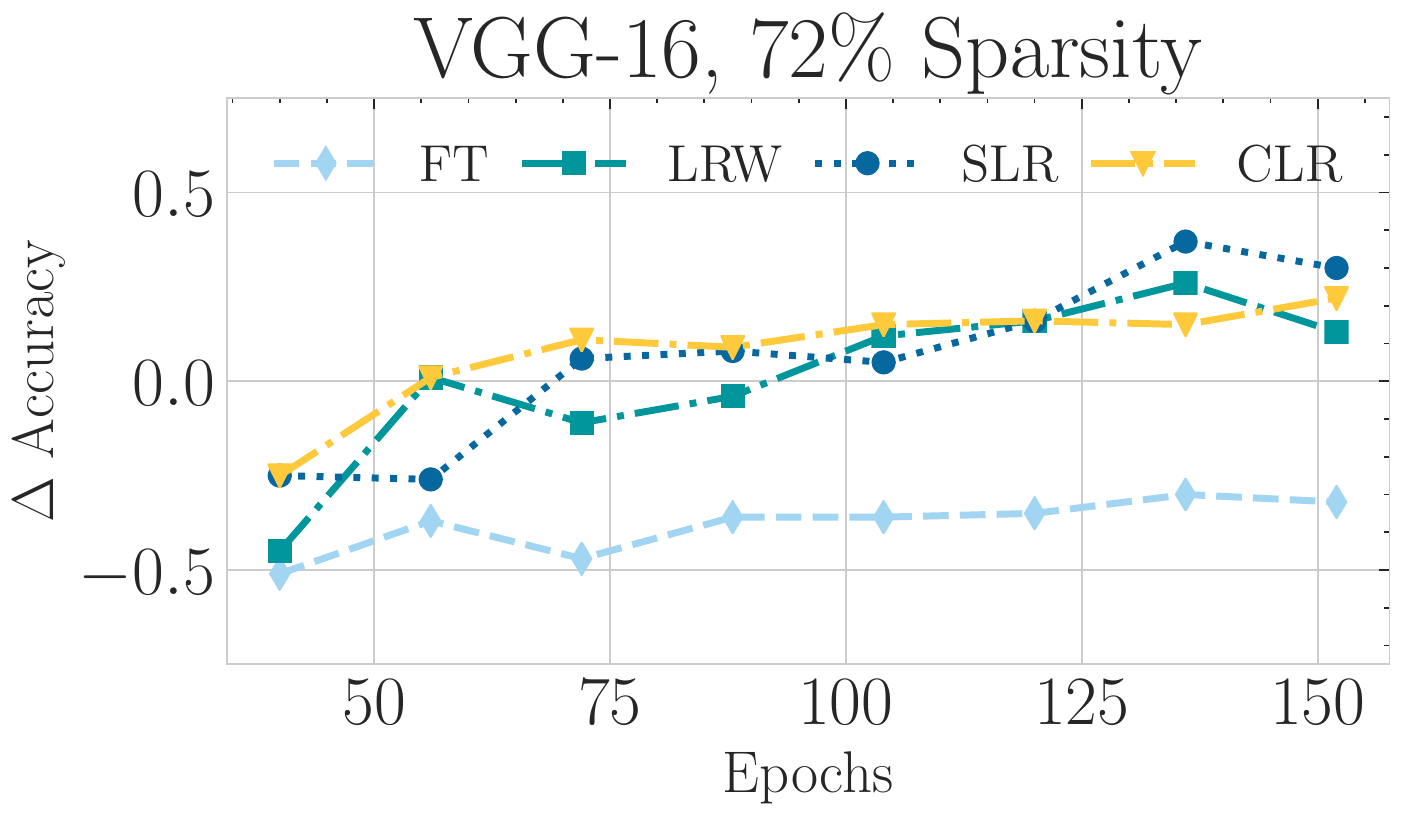}
			\label{fig:oneshot_pfec_cifar10_f}
		\end{subfigure}
		
		\caption{Results from pruning with high compression ratios for
			\textit{one-shot} \textit{structured} pruning on CIFAR-10 dataset using $\ell_1$-norm filters pruning \citep{li2016pruning}. With a proper learning rate schedule, it is possible to achieve almost no accuracy drop while having much more compact models and using slightly more training budgets.}
		\label{fig:oneshot_pfec_cifar10_compression}
	\end{figure}

	In addition results in Figure~\ref{fig:oneshot_pfec_cifar10_original}, we conduct experiments to examine impact of different retraining techniques to networks pruned with low and high compression ratios. The results are shown in Figure \ref{fig:oneshot_pfec_cifar10_compression}. 

	\begin{table}[t]
		\caption{Results (accuracy) for HRank ilters pruning \citep{lin2020hrank} on CIFAR-10. 
			``Pruned + FT$\dagger$'' is the model pruned from the large model with \emph{original} fine-tuning scheme that are reported. ``Prune + FT" and ``Prune + CLR"  are results of our runs with original retraining and CLR respectively. Configurations of Model and Pruned Model are both from the original paper. 
		}
		\small
		\centering
		\begin{tabular}{l|ccccccl}
			\toprule
			Model                       & Unpruned                                               & Prune + FT $\dagger$ & Prune + FT                                  & Prune + CLR   &  FLOPs $\downarrow\%$ & Params $\downarrow\%$  \\ \midrule
			VGG-16 (BN)                    & 93.93  & 92.34 & 91.97       & \textbf{92.53}     & $65.3\%$ & $82.10\%$\\  
			
			ResNet-56  & 93.26   &  93.17   & 92.97   & \textbf{93.16} & $50.0\%$ & $42.4\%$  \\ 
			
			ResNet-110 & 93.50 & 94.23 &  93.84 & \textbf{93.90} & $37.9\%$ & $38.7\%$ \\
			DenseNet-40 & 94.81 & 94.24 & 94.10 & 94.10 & $40.8\%$ & $30.5\%$ \\\bottomrule
			
		\end{tabular}
		\label{table:hrank-pruning}
	\end{table}
	
	\section{Evaluating CLR with Other Pruning Algorithms}
	\label{appendix:other_pruning}
	\textbf{Taylor Pruning} \citep{molchanov2019importance}. In this method, the authors estimated the contributions of each filter to the final performance by utilizing the first- and second-order Taylor expansion and performing iterative pruning. In the original implementation, the authors fine tuned the network with a small learning rate ($0.001$) for 25 epochs on ImageNet. 
	In our implementation, we opt to evaluate the accuracy of network retrained with learning rate restarting (CLR) with a relatively small budget. Therefore, we keep the retraining budget the same and only increase the learning rate to $0.01$.
	
	Table \ref{table:taylor-pruning} shows the accuracy of ResNet-50 on ImageNet when pruning $72\%$ of network weights. It can be seen that for a small retraining budget, employing learning rate restarting also gives significant gain in performance.
	
	\begin{table}[t]
		\caption{Results (accuracy) for Taylor filters pruning \citep{molchanov2019importance} on ImageNet. 
			All columns have same meaning with corresponding columns in Table \ref{table:hrank-pruning}.
		}
		\small
		\centering
		\begin{tabular}{c|c|cccc|}
			\toprule
			Model                       & Unpruned     & Prune + FT     & Prune + CLR   &  FLOPs $\downarrow\%$ & Params $\downarrow\%$  \\ \midrule
			
			Taylor-FO-BN-56\% & \multirow{3}{*}{76.15} & $71.69$ & $\mathbf{72.51}$ & $67.2$ & $66.8$ \\ 
			Taylor-FO-BN-72\%& & $74.50$ & $\mathbf{75.22}$ & $45.0$ & $44.5$  \\
			Taylor-FO-BN-81\%& & $75.48$ & $\mathbf{75.67}$ & $35.0$ & $30.1$ \\
			\bottomrule
		\end{tabular}
		\label{table:taylor-pruning}
	\end{table}
	
	\textbf{HRank} \citep{lin2020hrank}. Inspired by the discovery that average rank of feature maps of a filter is consistent, the authors suggested to iteratively remove low-rank feature maps that contain less information. Particularly,  we iteratively remove filters in each layer and then retrain the network for 30 epochs each time.  We examine the performance of pruned models with HRank  on CIFAR while employing {CLR} at each fine-tuning step. Due to the heavily expensive computational cost for retraining, we only run each model once for HRank. In the original version of this method, the authors used the learning rate of $0.01$ and drop by $10\times$ at epochs $5$ and $10$. In our implementation, we restart the learning rate to $0.1$ and apply the CLR schedule. 
	
	The quantitative results are shown in Table \ref{table:hrank-pruning}. In our experiments, {CLR} also attains comparable or better results than proposed retraining methods in original work.
	
	\section{Strong Baselines with $\ell_1$-norm Filters Pruning}
	\label{appendix:strong_baseline_with_pfec}
	\subsection{CIFAR-10}
	Furthermore, we show that by naively neglecting retraining configurations (i.e. budget), we can also reach state-of-the-art results on CIFAR-10 with PFEC~\cite{li2016pruning}. Specifically, we prune two ``standard" models namely ResNet-56 and ResNet-110 and retrain with CLR for 152 epochs, and then we compare with more sophisticated algorithms in Table \ref{table:pitfall_sota}.
	
	\begin{table}[!h]
		\caption{Results of ResNet-56 and ResNet-110 on CIFAR-10. The performance of other pruning algorithms are taken directly from original papers.
		}
		\small 
		\centering
		\begin{tabular}{c|l|ccccr}
			\toprule
			Model                       & Method &  Unpruned                                               &  Params $\downarrow$ & FLOPs $\downarrow$  & Prune & Acc. $\downarrow$ \\ \midrule
			\multirow{5}{*}{ResNet-56} & PFEC \citep{li2016pruning} & 93.04 & 14.1 & 27.6 & 93.06 & -0.02 \\
			& NISP \citep{yu2018nisp} & 93.26 & 42.4 & 35.5 & 93.01 & 0.25\\
			& HRank \citep{lin2020hrank} & 93.26 & 42.4 & 50.0 & 93.17 & 0.09\\
			& GAL \citep{lin2019towards} & 93.26 & - & 37.6 & 92.98 & 0.28\\
			& PFEC + CLR (ours) & 93.21 & 48.7 & - & 93.29 & -0.07\\
			\midrule
			\multirow{4}{*}{ResNet-110} & PFEC \citep{li2016pruning} & 93.53 &  32.6 & 38.7 & 93.30 & -0.23 \\
			& GAL-0.5 \citep{lin2019towards} & 93.50 & 44.8 & 48.4  & 92.74 & 0.76\\
			& HRank \citep{lin2020hrank} & 93.50 & 59.2 & 58.2  & 93.36 & 0.14\\
			& PFEC+CLR (ours) & 93.56 & 64.2 & - & 93.69 & -0.13\\
			\bottomrule
		\end{tabular}
		\label{table:pitfall_sota}
	\end{table}
	\subsection{ImageNet}
	\paragraph{Taylor Pruning \citep{molchanov2019importance}}  In this method, the authors estimated the contributions of each filter to the final performance by utilizing the first- and second-order Taylor expansion and performing iterative pruning.

	Original work retrains pruned networks for $25$ epochs with learning rate of $0.001$ with the exception for \textit{Taylor-FO-BN-56\%}. Table \ref{table:taylor_pfec} demonstrates the performance of  \textit{PFEC+CLR} compare with original works of \cite{molchanov2019importance}. In most cases, \textit{PFEC+CLR} attains higher or comparable accuracy with \textit{Taylor Pruning}.
	\begin{table}[!h]
		\caption{Comparing the performance of pruned network via PFEC + CLR and Taylor Pruning on ImageNet. The results of Taylor Pruning are taken directly from original papers.
		}
		\small 
		\centering
		\begin{tabular}{c|c|cc|cc}
			\toprule
			\textbf{Model}         & \textbf{Method}    &  \textbf{Param} $\downarrow\%$ & \textbf{FLOPs} $\downarrow\%$ &  \textbf{Unpruned Top-1}               &    \textbf{Top-1} \\ \midrule
			\multirow{2}{*}{ResNet-34} &    Taylor-FO-BN-82\% & $21.1$ & $22.3$ & $73.31$  &  $72.83$ \\
			&  PFEC + CLR  & $22.5$ & -  & $73.30$ & $\mathbf{73.01}$\\
			\midrule
			\multirow{8}{*}{ResNet-50}  & Taylor-FO-BN-56\%  & $66.8$ &   $67.2$ & $76.18$ & $\mathbf{71.69}$ \\ 
			& PFEC + CLR & $68.8$ & -  & $76.15$ & $70.70$ \\
			\cmidrule{2-6}
			& Taylor-FO-BN-72\% & $44.5$ & $45.0$ & 76.18 & $74.50$ \\
			& PFEC + CLR & $44.5$ &  & 76.15 & $\mathbf{75.04}$\\
			\cmidrule{2-6}
			& Taylor-FO-BN-81\% & $30.1$ & $35.0$  & 76.18 & $75.48$ \\
			& PFEC + CLR & $30.1$ & - & 76.15 & $\mathbf{75.79}$ \\
			\cmidrule{2-6}
			& Taylor-FO-BN-91\% & $11.4$ & $20.0$ & 76.18 & $\mathbf{76.43}$ \\
			& PFEC + CLR & $12.5$ & - & 76.15 & $76.37$ \\
			\bottomrule
		\end{tabular}
		\label{table:taylor_pfec}
	\end{table}
	
	\paragraph{Discrimination-aware Channel Pruning \citep{zhuang2018discrimination}} The authors introduce additional discrimination-aware losses into the network to increase the discriminative power of intermediate layers and then select the most discriminative channels for each layer by considering the additional loss and the reconstruction error.
	
	We use the retraining budget of $60$ epochs and set initial learning rate to $0.01$ same as \cite{zhuang2018discrimination}. However, we employ the 1-cycle learning rate instead of stepwise learning rate as original work. The detailed comparison is presented in Table \ref{table:dcp_pfec}.
	
	\begin{table}[!h]
		\caption{Comparing the performance of pruned network via PFEC + CLR and DCP on ImageNet. The results of DCP are taken directly from original papers.
		}
		\small 
		\centering
		\begin{tabular}{c|cc|c|cc}
			\toprule
			\textbf{Model}                       & \textbf{Param $\downarrow\%$} & \textbf{FLOPs $\downarrow\%$} &  \textbf{Method}                                               &  \textbf{Unpruned Top-1} & \textbf{Top-1} \\ \midrule
			\multirow{6}{*}{ResNet-18} & 28.1 & 27.1 &  DCP & $69.64$ & $69.21$ \\
			& 31.9 & - & PFEC + CLR  & $69.76$ & $\mathbf{69.31\pm0.06}$ \\
			\cmidrule{2-6}
			& 47.1 & 46.1 & DCP  & $69.64$ & $67.35$ \\
			& 50.6 & - & PFEC + CLR & $69.76$ & $\mathbf{67.38}$\\
			\cmidrule{2-6}
			& 65.7 & 64.1 & DCP & $69.64$ & $\mathbf{64.12}$ \\
			& - & - & PFEC + CLR & $69.76$ & $64.08$ \\
			\midrule
			
			\multirow{6}{*}{ResNet-50} & 33.3 & 35.7 & DCP & $76.01$ & $\mathbf{76.40}$\\
			& 33.7 & - & PFEC  + CLR & $76.15$ & $76.03$ \\
			\cmidrule{2-6}
			& 51.4 & 55.5 & DCP & $76.01$ & $74.95$ \\
			& 51.5 & - & PFEC + CLR & $76.15$ &  $\mathbf{75.16}$ \\
			\cmidrule{2-6}
			& 65.9 & 71.1 & DCP & $76.01$ & $72.75$\\
			& 66.1 & - & PFEC + CLR & $76.15$ & $\mathbf{72.92}$ \\
			\bottomrule
		\end{tabular}
		\label{table:dcp_pfec}
	\end{table}
	\paragraph{Provable Filters Pruning \citep{Liebenwein2020Provable}} This algorithm uses a small batch of input data points to assign a saliency score to each filter and constructs an importance sampling distribution where filters that highly affect the output are sampled with correspondingly high probability.
	
	
	Original implementation of \cite{Liebenwein2020Provable} retrain the pruned network on ImageNet for $90$ epochs with standard learning rate schedule (drop learning rate by factor of $10$ after 30 epochs) with batch size of $256$ and weight decay of $0.0001$. In our implementation, pruned networks are trained for $90$ epochs with maximum learning rate of $0.1$ while keeping the same configurations for all other hyperparameters as original work.
	
	\begin{table}[!h]
		\caption{Comparing the performance of pruned network via PFEC + CLR and Provable Filters Pruning (PFP) on ImageNet. The results of PFP are taken directly from original paper \citep{Liebenwein2020Provable}. 
		}
		\small 
		\centering
		\begin{tabular}{c|c|cc|cc}
			\toprule
			\textbf{Model}        & \textbf{Method}    &  \textbf{Param} $\downarrow\%$ & \textbf{FLOPs} $\downarrow\%$    &  \textbf{Unpruned Top-1}              &    \textbf{Top-1} \\ \midrule
			\multirow{4}{*}{ResNet-50} & PFP (lowest top-1 err.) & $18.0$ & $10.8$  & $76.13$ &  $75.91$ \\
			&  PFEC + CLR & $18.1$ & -  &   $76.15$  &  $\mathbf{76.56}$ \\
			
			\cmidrule{2-6}
			& PFP (within $1.0\%$ top-1)  & $44.0$ & $30.1$  & $76.13$ &  $75.21$\\
			& PFEC + CLR  & $44.3$ & -  & $76.15$ & $\mathbf{75.38}$\\
			
			\bottomrule
		\end{tabular}
		\label{table:hrankplus_pfec}
	\end{table}
	\section{Random Pruning with CLR}
	\subsection{Retraining Configuration}
	\label{appendix:original_random_pruning}
	We compare the configuration between original implementation of aforementioned pruning methods and corresponding random pruning in Table \ref{table:random_pruning_configuration}. Note that for HRankPlus, the original implementation employs different values of weight decay for each model: $0.005, 0.006, 0.005, 0.002$ for VGG-16, ResNet-56, ResNet-110, DenseNet-40 respectively. We found that these value is relatively higher than ``conventional" values (0.0001) making the models hard to converge after restarting. Thus, we use weight decay of $0.0005$ through out experiments with HRankPlus. In all our experiments, other details such as batch size, retraining budget are set to similar value of original implementation.
	\begin{table}[!h]
		\caption{Training configurations of original pruning methods and random pruning.
		}
		\small 
		\centering
		\begin{tabular}{c|c|c|c}
			\toprule
			\textbf{Method} & \textbf{Model} & \textbf{Original} & \textbf{Random Pruning} \\
			\midrule
			\multirow{12}{*}{HRankPlus} & \multirow{2}{*}{ResNet-56} & \multirow{6}{*}{$\alpha = \begin{cases} 0.01 & \text{t}\in [0, 150) \\ 0.001 & \text{t}\in [151, 225)\\ 0.0001 & \text{t}\in [226, 300)\end{cases}$} & \multirow{12}{*}{$\begin{cases}\alpha_{\text{init}} = 0.001 \\\alpha_{\text{max}}=0.1 \\ \alpha_{\text{min}}=0.00001\end{cases}$}\\
			& & & \\
			& \multirow{2}{*}{ResNet-110} & & \\
			& & & \\
			& \multirow{2}{*}{DenseNet-40} & & \\
			& & & \\
			\cmidrule{2-3}
			& \multirow{6}{*}{VGG-16} & \multirow{6}{*}{$\alpha = \begin{cases} 0.01 & \text{t}\in [0, 50) \\ 0.001 & \text{t}\in [51, 100)\\ 0.0001 & \text{t}\in [101, 150)\end{cases}$} & \\
			& & & \\
			& & & \\
			& & & \\
			& & & \\
			& & & \\
			\cmidrule{1-4}
			\multirow{3}{*}{PFEC} & \multirow{3}{*}{ResNet-34} & \multirow{3}{*}{$\alpha=0.001 \hspace{5mm} \text{t}\in[0, 20)$} & \multirow{6}{*}{$\begin{cases}\alpha_{\text{init}} = 0.0001 \\\alpha_{\text{max}}=0.01 \\ \alpha_{\text{min}}=0.000001\end{cases}$}\\
			& & & \\
			& & & \\
			\cmidrule{1-3}
			\multirow{3}{*}{Taylor Pruning} & \multirow{3}{*}{ResNet-50} & \multirow{3}{*}{$\alpha=0.001 \hspace{5mm} \text{t}\in[0, 25)$} & \\
			& & & \\
			& & & \\
			\bottomrule
		\end{tabular}
		\label{table:random_pruning_configuration}
	\end{table}
	\subsection{Random Pruning with Various Compression Ratio}
	\label{appendix:random_pruning_compression_ratio}
\begin{figure}
		\centering
		\def\sc{0.245}
		\begin{subfigure}[b]{\sc\textwidth}
			\centering
			\includegraphics[width=\textwidth]{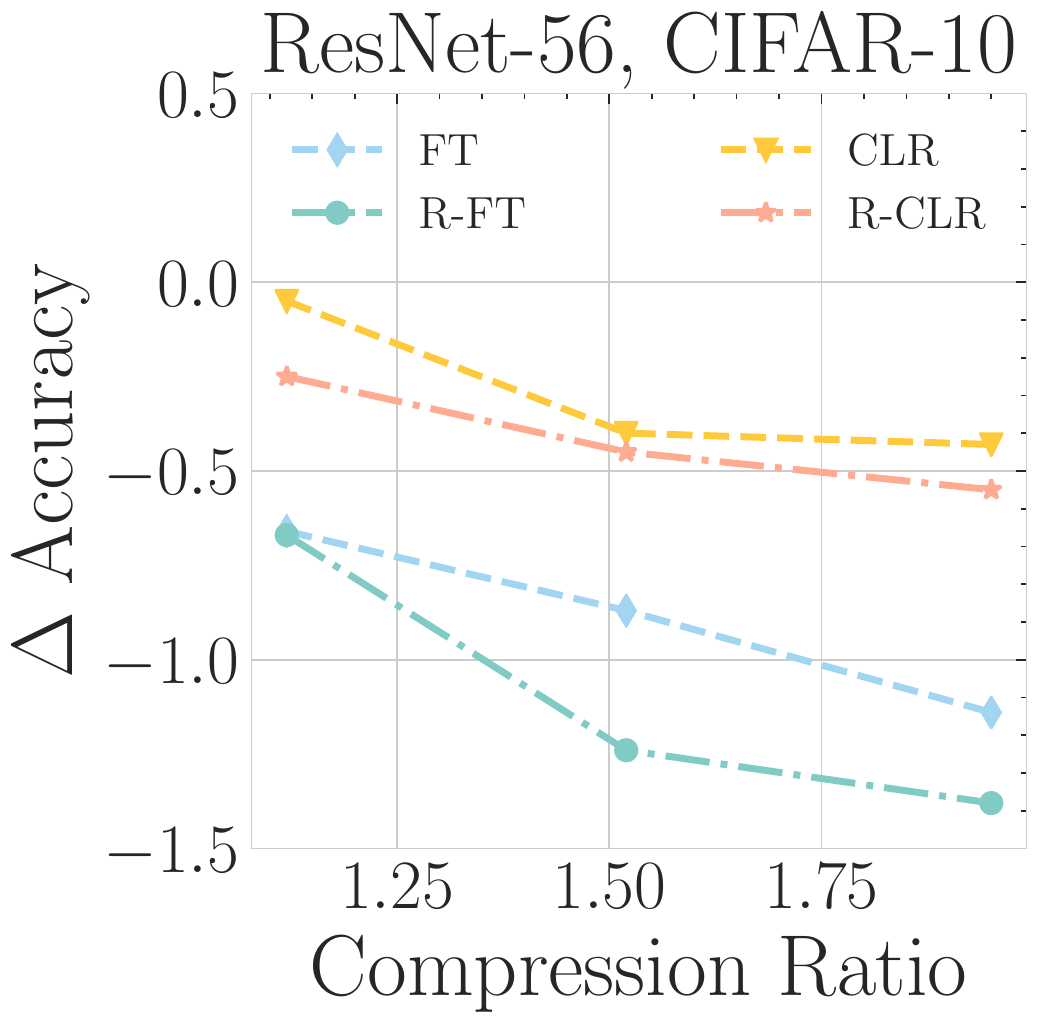}
			\caption{ResNet-56}
			\centerline{\small CIFAR-10}
			\label{fig:random_cifar_a}
		\end{subfigure}
		\hfill
		\begin{subfigure}[b]{\sc\textwidth}
			\centering
			\includegraphics[width=\textwidth]{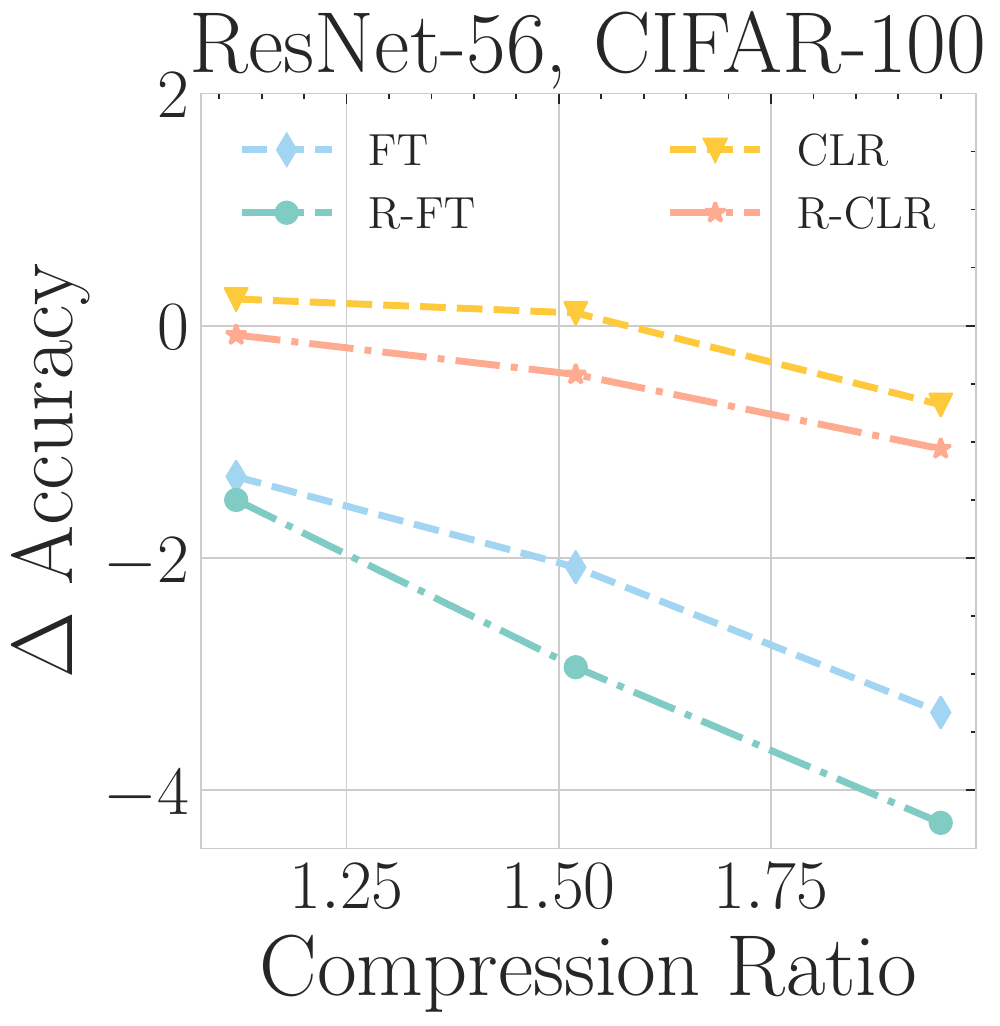}
			\caption{ResNet-56}
			\centerline{\small CIFAR-100}
			\label{fig:random_cifar_b}
		\end{subfigure}
		\hfill
		\begin{subfigure}[b]{\sc\textwidth}
			\centering
			\includegraphics[width=\textwidth]{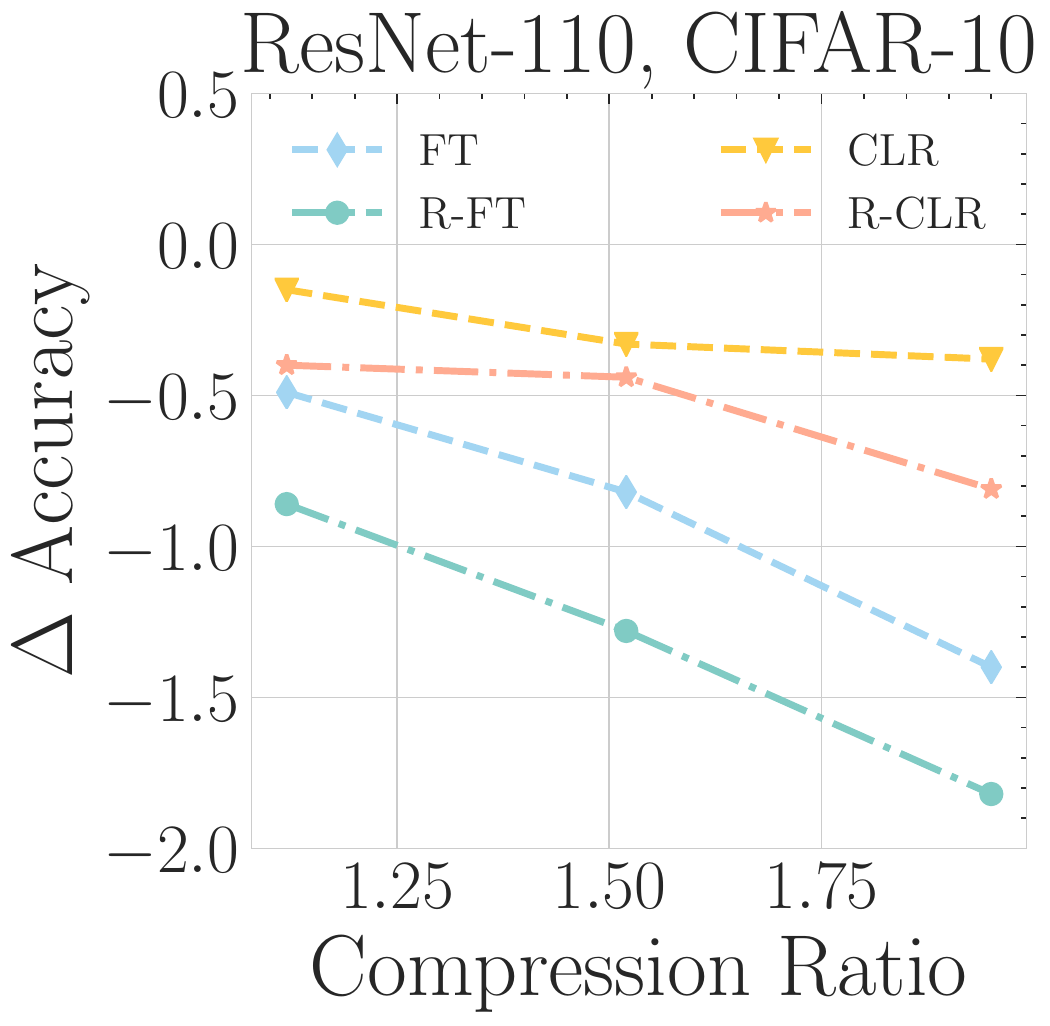}
			\caption{ResNet-110}
			\centerline{\small CIFAR-10}
			\label{fig:random_cifar_c}
		\end{subfigure}
		\hfill
		\begin{subfigure}[b]{\sc\textwidth}
			\centering
			\includegraphics[width=\textwidth]{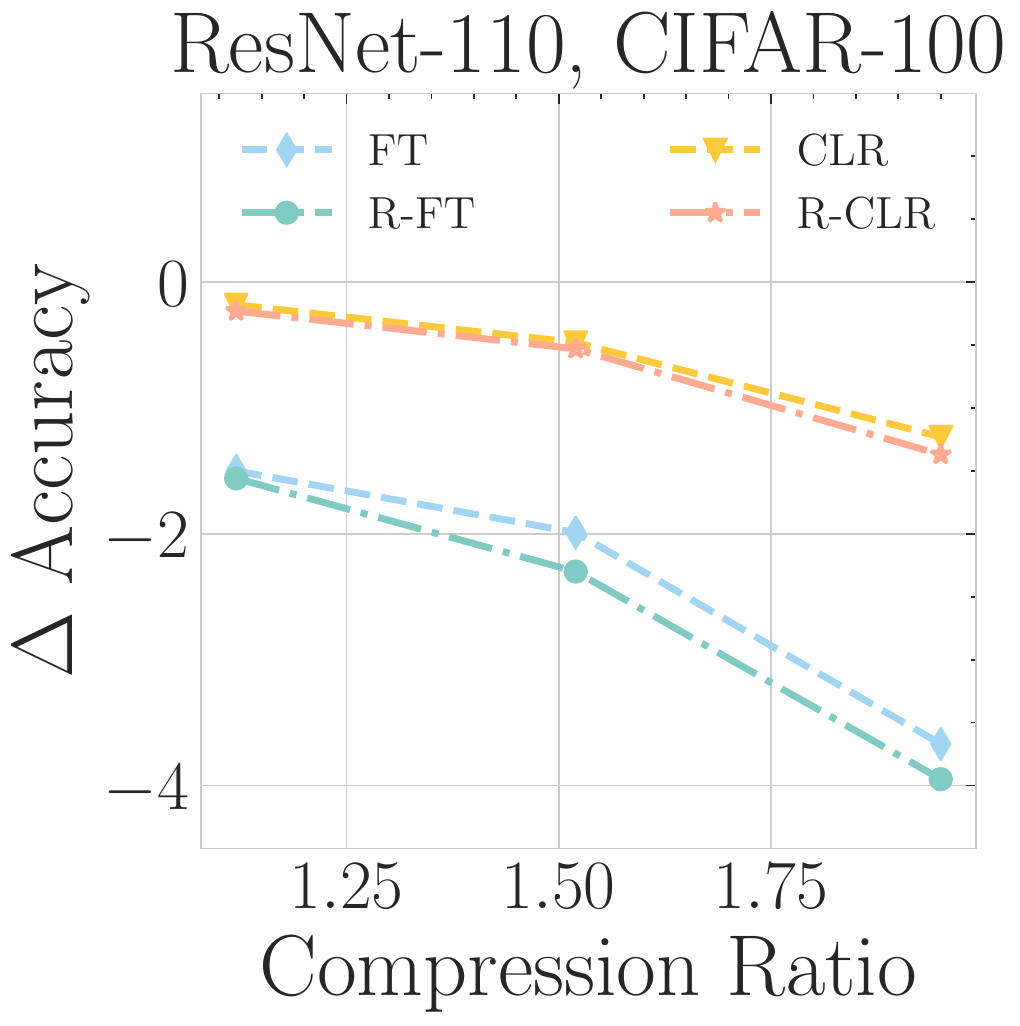}
			\caption{ResNet-110}
			\centerline{\small CIFAR-100}
			\label{fig:random_cifar_d}
		\end{subfigure}
		\caption{\textit{One-shot} \textit{structured} pruning on CIFAR-10 dataset using PFEC \citep{li2016pruning} and randomly filters pruning with different retraining schemes.}
		\label{fig:random_cifar}
	\end{figure}

	\paragraph{$\ell_1$-norm Filters Pruning} Figure \ref{fig:random_cifar} illustrates the performance of networks pruned with PFEC and random pruning on CIFAR-10 and CIFAR-100 when retraining with $40$ epochs -- the same setting used by \cite{li2016pruning}. We can see that randomly pruned networks consistently achieve \textit{superior} performance than methodically pruned networks (fine-tuned with standard learning rate schedule) in terms of accuracy. However, random pruning obtain lower accuracy than PFEC when using identical retraining techniques.
	
\begin{figure}[t]
		\centering
		\def\sc{0.49}
		\begin{subfigure}[b]{\sc\textwidth}
			\centering
			\includegraphics[width=\textwidth]{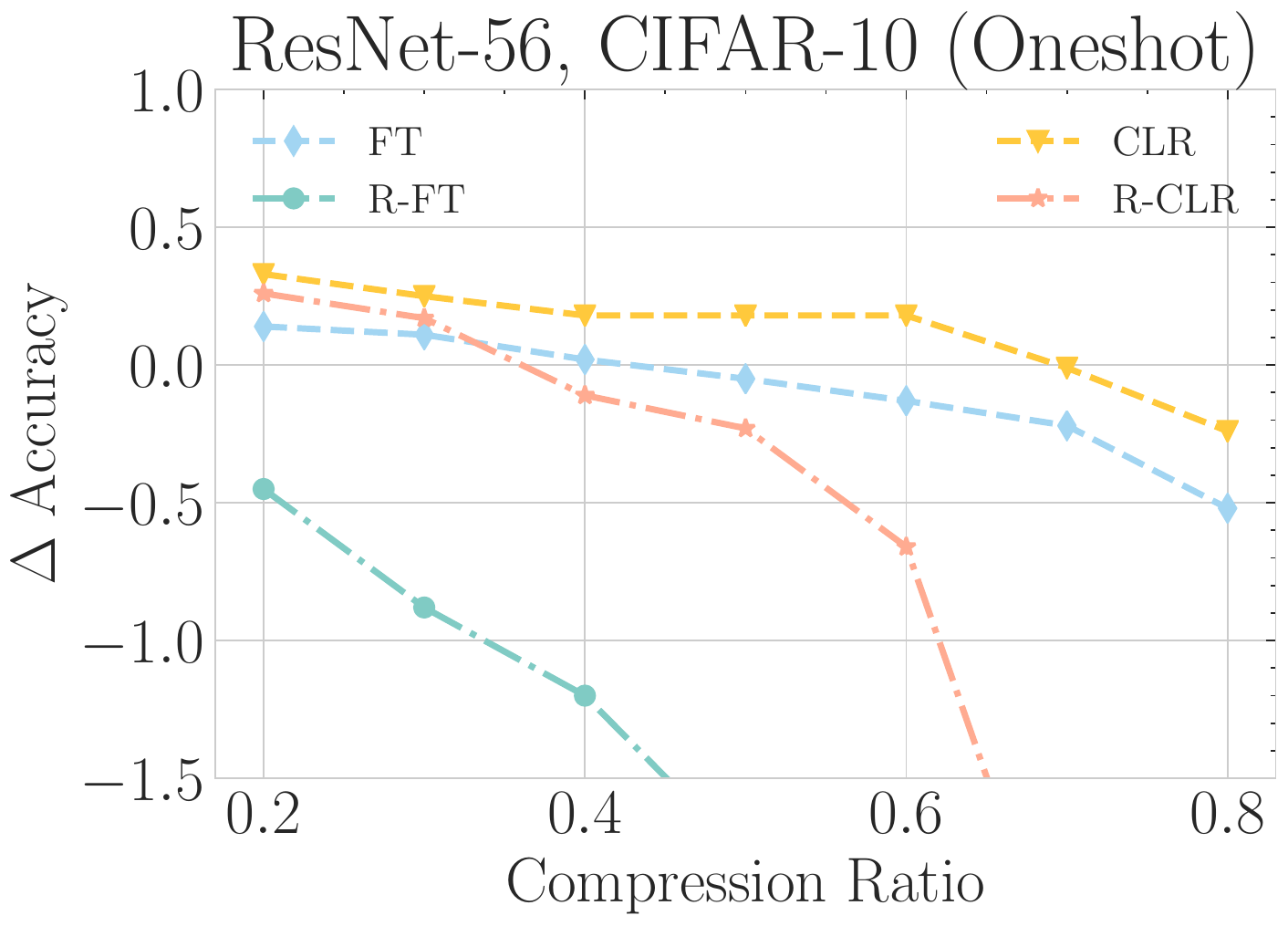}
			\label{fig:resnet56_mwp_random_a}
		\end{subfigure}
		\hfill
		\begin{subfigure}[b]{\sc\textwidth}
			\centering
			\includegraphics[width=\textwidth]{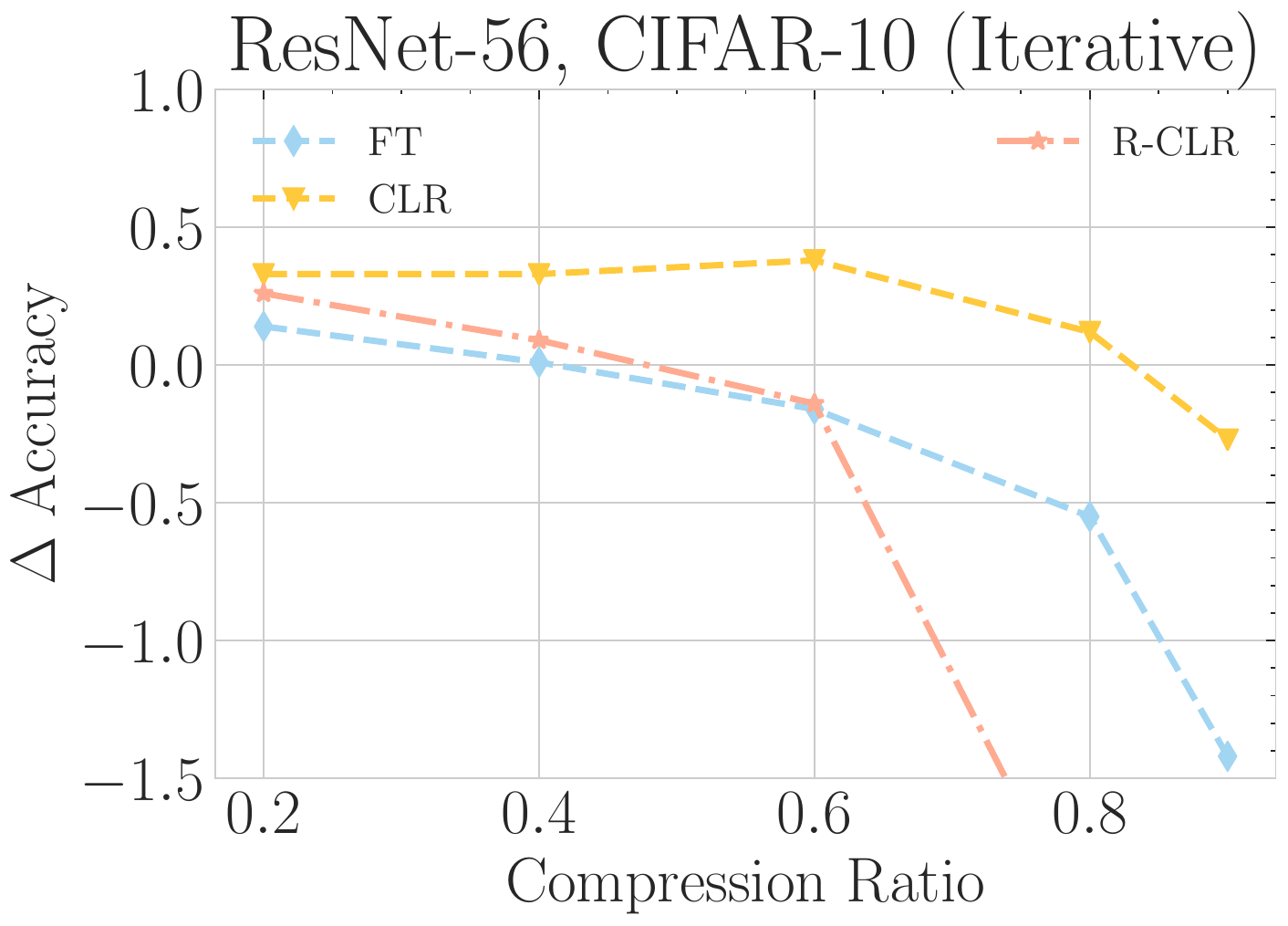}
			\label{fig:resnet56_mwp_random_b}
		\end{subfigure}
		
		\caption{\textit{One-shot} and \textit{Iterative} \textit{unstructured} pruning on CIFAR-10 dataset using MWP \citep{han2015learning} and randomly weights pruning with different retraining schemes.}
		\label{fig:resnet56_mwp_random}
	\end{figure}

	\paragraph{Magnitude-based Weights Pruning} We extend the scope of experiment in Sec \ref{sec:random_pruning} to \emph{unstructured} pruning and analyze performance of MWP with CLR. Figure \ref{fig:resnet56_mwp_random} represents the results of random pruning with CLR and methodically pruning while varying difference compression ratio in both iterative and oneshot pruning manner. Specifically, we retrain the trimmed network for $40$ epochs. We can observe that though Random Pruning + CLR can achieve higher accuracy with low sparsity, the performance of randomly pruned network immensely reduced with the increasing of compression ratio.
	
	\section{Performance of CLR with Different Warmup Epochs}
	\label{appendix:clr_warmup}
	To avoid the laborious hyperparameters tuning process, we simply warm up the learning rate for CLR and SLR for the first $10\%$ budget of retraining. In this section, we investigate whether the performance of pruned network is sensitive to this hyperparameter.
	
\begin{figure}
		\centering
		\includegraphics[width=0.75\textwidth]{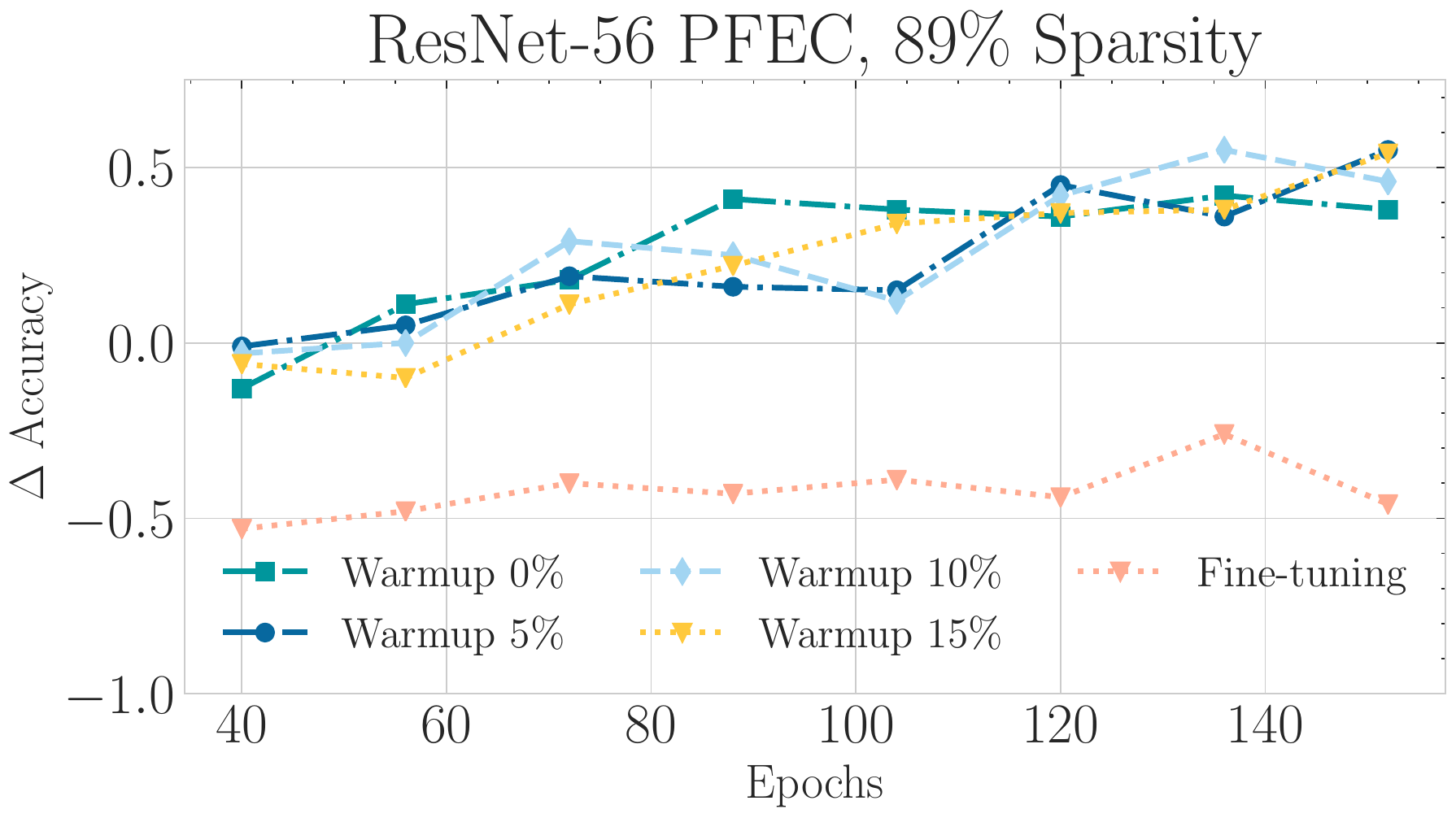}
		\caption{$\ell_1$-norm Filters Pruning \citep{li2016pruning} on CIFAR-10 with different number of warmup epochs.}
		\label{fig:warmup}
	\end{figure}

	Figure \ref{fig:warmup} presents the results of pruned networks obtained with fine-tuning and CLR while varying number of warmup epochs namely $0, 5, 10, 15\%$ of total retraining. We can observe that there is no significant diversity in results of CLR with different number of warmup epochs and all of them exceed performance of fine-tuning by a large margin.

	\section{Training from Scratch with CLR}
	In this section, we conduct experiment where we train the unpruned models from scratch with CLR schedule. Then, we compare performance of networks pruned from these baseline with those pruned from models trained with conventional step wise learning rate schedule. Table \ref{table:training_from_scratch_clr} presents the accuracy of pruned ResNet-56 models on CIFAR-10 with different training and retraining recipes. We use the budget of $160$ for training the baseline networks. For retraining the trimmed networks, we examine the budget of $40$ and $56$ epochs. We can see that both CLR and SLR consistently exceed the performance of fine-tuning under these configurations.
	\begin{table}[t]
		\caption{Performance of ResNet-56 pruned from models trained with CLR and conventional step-wise learning rate schedule on CIFAR-10. ``Epochs" column indicates the number of epochs for retraining trimmed networks. ``Schedule" colum indicates the learning rate schedule used for training baseline (unpruned) networks. The best and second best methods are highlighted in \textbf{\color{blue}{bold blue}} and {\color{blue}blue} respectively.}
		\small 
		\centering
		\begin{tabular}{c|c|cH|c|ccc}
			\toprule
			\textbf{Epochs} & \textbf{Schedule}    &  \textbf{Param} $\downarrow\%$ & \textbf{FLOPs} $\downarrow\%$    &  \textbf{Baseline}              &    \textbf{Fine-tuning} & \textbf{SLR} &  \textbf{CLR} \\ \midrule
			\multirow{4}{*}{40} & Step-wise & $13.7$ & $10.8$  & $93.15\pm0.36$ & $92.81\pm0.49$ & $\secondbest{92.93\pm0.15}$ & $\best{93.14\pm0.37}$\\
			& CLR & $13.7$ & -  &   $93.45\pm0.17$  &  $\secondbest{93.06\pm0.20}$ & $93.03\pm0.16$ & $\best{93.15\pm0.11}$\\
			\cmidrule{2-8}
			& Step-wise & $34.9$ & $10.8$  & $93.15\pm0.36$ & $92.25\pm0.35$ & $\best{92.83\pm0.16}$  & $\secondbest{92.81\pm0.32}$ \\
			&CLR & $34.9$ & -  &   $93.45\pm0.17$  &  $92.50\pm0.07$ & $\secondbest{92.63\pm0.05}$ & $\best{93.03\pm0.16}$\\
			
			\midrule 
			
			\multirow{4}{*}{56} & Step-wise & $13.7$ & $10.8$  & $93.15\pm0.36$ & $92.81\pm0.41$ & $\secondbest{92.86\pm0.28}$ & $\best{93.22\pm0.37}$ \\
			& CLR & $13.7$ & -  &   $93.45\pm0.17$  &  $93.04\pm0.16$ & $\secondbest{93.22\pm0.05}$ & $\best{93.37\pm0.25}$ \\
			\cmidrule{2-8}
			& Step-wise & $34.9$ & $10.8$  & $93.15\pm0.36$ & $92.48\pm0.45$ & $\secondbest{92.89\pm0.11}$ & $\best{93.26\pm0.17}$\\
			&CLR & $34.9$ & -  &   $93.45\pm0.17$  &  $92.63\pm0.13$ & $\secondbest{92.83\pm0.03}$ & $\best{93.29\pm0.25}$ \\
			
			\bottomrule
		\end{tabular}
		\label{table:training_from_scratch_clr}
	\end{table}

\end{document}